\documentclass[journal]{IEEEtran}

\usepackage{cite}
\usepackage{booktabs}
\usepackage{multirow}
\usepackage{array}
\usepackage{graphicx}
\usepackage{amsmath,amssymb,bm}
\usepackage[table]{xcolor}
\usepackage{siunitx}
\usepackage{enumitem}
\usepackage{microtype}
\usepackage{xspace}
\usepackage{url}
\usepackage[hidelinks]{hyperref}
\usepackage{cleveref}
\usepackage{pifont}
\usepackage{tabularx}

\newcommand{\smep}{SMM Planner\xspace}
\newcommand{\vtep}{MSTS-Trained Planner\xspace}
\newcounter{algorithm}
\crefname{algorithm}{Algorithm}{Algorithms}
\Crefname{algorithm}{Algorithm}{Algorithms}
\newenvironment{ruledalgorithm}[1]{%
  \refstepcounter{algorithm}%
  \par\vspace{2pt}%
  \hrule height 0.5pt
  \vspace{2pt}%
  \noindent\textbf{Algorithm~\thealgorithm:} #1\par
  \vspace{2pt}%
  \hrule height 0.5pt
  \vspace{3pt}%
  \footnotesize
  \setlength{\abovedisplayskip}{3pt}%
  \setlength{\belowdisplayskip}{3pt}%
  \setlength{\jot}{1pt}%
}{%
  \vspace{1pt}%
  \hrule height 0.5pt
  \par\vspace{2pt}%
}

\begin{document}
\bstctlcite{stabledrive_bst_control}

\title{Not All History Helps: Velocity-Aware Selective Memory for Long-Horizon End-to-End Autonomous Driving}

\author{Yuchen Liu, Ziying Song, Shengkai Zhang, Jiannan Chen, Peiliang Wu, Lei Yang, Bin Sun, Yan Gong, and Li Wang%
\thanks{This work was supported by the National Natural Science Foundation of China under Grant No. 52502496, the Beijing Natural Science Foundation under Grant No. L2609087, and the Natural Science Foundation of Chongqing, China under Grant No. CSTB2025NSC0-GPX0413. (Corresponding author: Ziying Song.)}%
\thanks{Yuchen Liu is with Nanyang Technological University, Singapore, and North University of China, Taiyuan 030051, China.}%
\thanks{Ziying Song is with Nanyang Technological University, Singapore, and Yanshan University, Qinhuangdao 066004, China (e-mail: songziying@ysu.edu.cn).}%
\thanks{Jiannan Chen and Peiliang Wu are with Yanshan University, Qinhuangdao 066004, China.}%
\thanks{Shengkai Zhang is with Beijing Jiaotong University, Beijing 100044, China.}%
\thanks{Lei Yang is with Nanyang Technological University, Singapore.}%
\thanks{Bin Sun is with China Automotive Technology and Research Center Co., Ltd., Beijing 100191, China.}%
\thanks{Yan Gong is with Harbin Institute of Technology, Harbin 150001, China.}%
\thanks{Li Wang is with Beijing Institute of Technology, Beijing 100081, China.}%
}

\maketitle

\begin{abstract} Reliable long-horizon planning remains a key challenge in end-to-end autonomous driving. By accounting for future motion evolution and potential consequences, it provides forward-looking guidance for safe and consistent driving in evolving traffic environments. Existing methods use historical planning states as temporal context. Self-generated history may become stale or conflict with the current motion stage, introducing unreliable priors. We propose StableDrive to address cross-cycle historical reliability and within-horizon motion-stage evolution. Selective Momentum Memory (SMM), implemented with a Mamba selective state-space operator, controls the influence of the preceding self-predicted planning state on the current cycle. Motion-Stage Training Scaffold (MSTS) uses motion-stage, long-horizon trajectory, and longitudinal-motion supervision to guide stage-aware future motion learning and is removed before inference. A fixed parameter midpoint between two architecture-aligned endpoints yields a single deployable SMM planner without model ensembling or extra inference-time computation. On nuScenes under the MomAD evaluation protocol, StableDrive achieves SOTA performance across all reported planning metrics from 1 to 6 s, reducing average collision rate by 23.3\%, TPC by 30.9\%, and L2 by 11.8\% over the best previously reported value for each metric. On the curated Longitudinal-Transition nuScenes (LT-nuScenes), StableDrive reduces 6-s collision rate by 23.81\%, TPC by 10.90\%, and L2 by 6.37\%. On NAVSIM v1 and v2, StableDrive achieves the highest PDMS/EPDMS in all three reported settings, including a 5.7-point EPDMS gain on v2 navhard over the previous best. \end{abstract}


\IEEEpeerreviewmaketitle

\section{Introduction}
\label{sec:intro}

\IEEEPARstart{E}{nd-to-end} autonomous driving (E2E-AD) unifies perception, prediction, and planning within a single learning framework and optimizes scene representations toward the final driving objective. Compared with conventional modular pipelines, this planning-oriented paradigm reduces information fragmentation across individual tasks and establishes a more direct connection between scene understanding, motion prediction, and ego-vehicle decision-making~\cite{hu2023uniad,jiang2023vad,li2024sparsedrive,tits_dong2026_monocular,tits_yin2026_navdrive,tits_kang2026_planning}.

Unified modeling does not naturally guarantee reliable decisions over extended horizons~\cite{song2025momad,graphworld2026,tits_shao2025_uncertainty}. The practical performance of an end-to-end driving system ultimately depends on whether its planner can understand how the scene evolves over time and generate accurate and coherent future trajectories accordingly. This challenge is particularly pronounced in long-horizon planning. Long-horizon planning is not merely a matter of appending additional waypoints to a short-term trajectory~\cite{zhang2025bridgead,graphworld2026}. Instead, the planner must anticipate how the ego vehicle will transition through different motion phases, including acceleration, cruising, braking, stopping, restarting, and car following. As the planning horizon increases, small errors in velocity, acceleration, and following distance can accumulate over time, eventually leading to substantial trajectory deviations, delayed braking, inaccurate stopping positions, and increased collision risk~\cite{song2025momad,graphworld2026,tits_shao2025_uncertainty}. Therefore, the difficulty of long-horizon planning lies not only in predicting farther into the future, but also in maintaining reliable temporal reasoning as the vehicle motion state continuously evolves.

Existing methods use historical and temporal context to improve long-horizon planning. MomAD~\cite{song2025momad} introduces trajectory and perception momentum to refine the current plan with historical planning queries. BridgeAD~\cite{zhang2025bridgead} aligns historical prediction and planning queries with individual future timesteps. GraphWorld~\cite{graphworld2026} models future scene evolution through latent world states and interactions among traffic participants. These methods show the value of temporal context but leave a central question largely unresolved: \emph{when should self-generated historical planning states continue to be trusted?}

\begin{figure*}[t]
  \centering
  \includegraphics[width=0.97\textwidth]{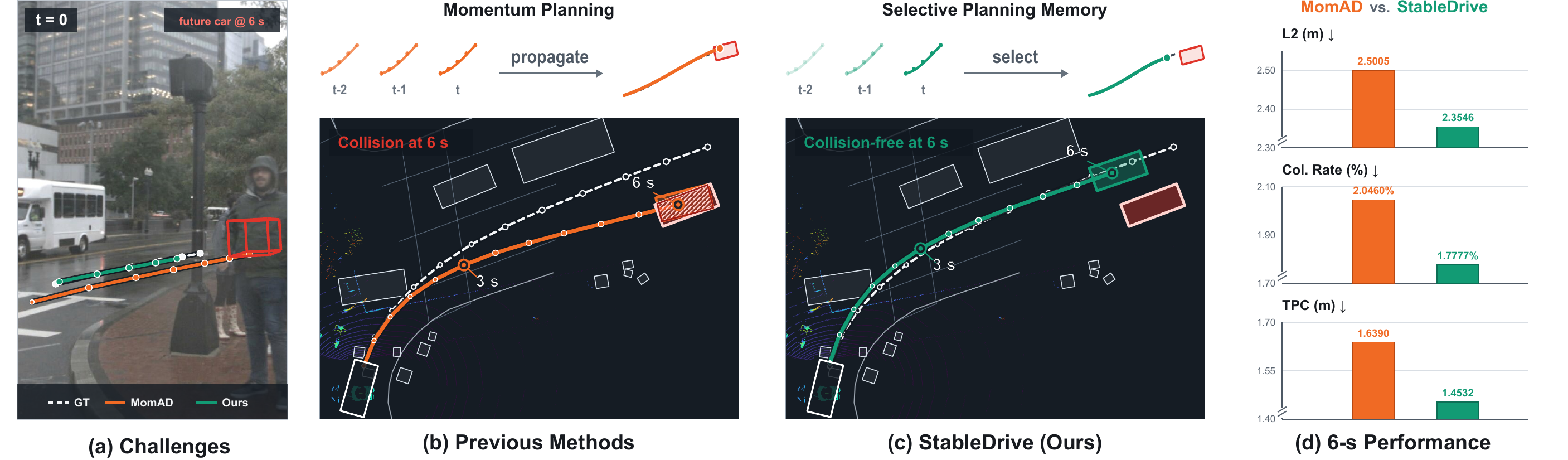}
  \caption{\textbf{Selective planning memory enables safer long-horizon planning.} (a) A nuScenes example comparing MomAD~\cite{song2025momad} and StableDrive with the ground-truth future trajectory. (b) Retaining a stale planning prior causes MomAD~\cite{song2025momad} to diverge after \SI{3}{s} and collide at \SI{6}{s}. (c) StableDrive suppresses unreliable history and remains collision-free. (d) StableDrive improves six-second L2, Col. Rate, and TPC by 5.83\%, 13.11\%, and 11.34\%, respectively, over our local MomAD~\cite{song2025momad} reproduction under the same evaluation protocol. Lower is better for all metrics.}
  \label{fig:motivation}
\end{figure*}

This issue becomes more pronounced as the planning horizon extends. \Cref{fig:motivation} shows a representative case from nuScenes~\cite{caesar2020nuscenes}. At \SI{3}{s}, MomAD~\cite{song2025momad} and StableDrive produce trajectories that are similarly close to the ground truth. Beyond this point, MomAD~\cite{song2025momad} continues to carry forward a stale planning prior, and its trajectory gradually diverges from the ground truth before overlapping with the future occupancy of another vehicle at \SI{6}{s}. StableDrive instead suppresses the unreliable history and remains collision-free. The same trend appears in the aggregate results: under the same evaluation protocol, StableDrive achieves lower 6-s L2, Col. Rate, and TPC than our reproduced MomAD~\cite{song2025momad} baseline. This pattern is consistent with a key property of planning history: unlike direct sensor observations, historical planning states are predictions generated in earlier planning cycles~\cite{song2025momad,zhang2025bridgead} and therefore retain both estimation errors and previous motion assumptions. They can provide useful temporal cues during stable motion, but their longitudinal trends may become invalid when the ego vehicle brakes, stops, restarts, or establishes a new car-following relationship. In such cases, the lateral geometry may remain plausible even though the longitudinal prior has already become stale. The comparison in \Cref{sec:h4} provides further supporting evidence: the one-frame StableDrive setting achieves lower L2 and Col. Rate at every reported horizon and a lower average TPC than the corresponding two- and four-frame settings. The key issue is therefore not how much history is retained, but whether it remains reliable for the current planning cycle.

Selective state space models provide a suitable foundation for addressing this problem. Mamba~\cite{gu2023mamba} performs input-dependent state updates, allowing information to be selectively retained or suppressed during sequence modeling, while Mamba-2~\cite{dao2024mamba2} further develops this formulation through structured state space duality. Recent autonomous-driving applications include Trajectory Mamba~\cite{huang2025trajectorymamba} and MLSTP~\cite{tits_ren2026_mlstp} for motion forecasting, MamBEV~\cite{ke2025mambev} for BEV representation, GMF-Drive~\cite{wang2025gmfdrive} for spatial feature fusion, and DriveMamba~\cite{drivemamba2026} for unified end-to-end decoding. However, these applications do not explicitly address whether self-generated planning states remain reliable after longitudinal motion-stage transitions or when stale history should be suppressed. Moreover, a generic Mamba~\cite{gu2023mamba} module does not define the source of cross-cycle planning memory, its causal boundary, or its relation to ego-planning semantics. Reliable long-horizon planning therefore requires a selective memory mechanism specifically designed for historical ego-planning states.

Based on these observations, we propose \textbf{StableDrive}, a reliable long-horizon end-to-end planning framework that addresses temporal uncertainty at two complementary scales. \textbf{Selective Momentum Memory (SMM)} uses a Mamba selective state-space operator~\cite{gu2023mamba} to establish a causal memory pathway across consecutive planning cycles and regulate how strongly the preceding self-prediction influences the current query. Since its cache is formed solely from predictions generated in previous planning cycles, SMM requires neither future information nor ground-truth planning history. \textbf{Motion-Stage Training Scaffold (MSTS)} uses motion-stage, long-horizon trajectory, and longitudinal-motion supervision to guid

e the shared planner in learning within-horizon motion-stage evolution. MSTS is removed through checkpoint construction before inference, while the effect of MSTS-assisted optimization remains in the retained shared parameters. The two training paths produce architecture-aligned endpoints with complementary empirical behavior: the MSTS-assisted endpoint provides substantial gains in trajectory accuracy and slightly better temporal consistency, whereas the SMM-only endpoint yields larger gains in collision safety. We therefore consolidate them through \textbf{fixed midpoint checkpoint construction}, resulting in a single deployable SMM checkpoint and one forward path without model ensembling or additional inference-time computation.

Extensive experiments on nuScenes demonstrate the effectiveness of this design. Across the 4--6~s planning horizons, StableDrive achieves state-of-the-art performance across all nine reported metrics under the same evaluation protocol, reducing L2 by \textbf{4.59\%--5.83\%}, Col. Rate by \textbf{13.11\%--27.10\%}, and TPC by \textbf{9.51\%--11.34\%}. Further component-wise ablations and endpoint analyses verify the roles of SMM, MSTS, and fixed midpoint checkpoint construction. These results show that reliable long-horizon planning does not depend on indiscriminately extending temporal context. Instead, it requires controlled cross-cycle historical influence together with within-horizon motion-stage guidance.

Our main contributions are summarized as follows:

\begin{itemize}[leftmargin=*]
    \item We propose \textbf{StableDrive}, a reliable long-horizon end-to-end planning framework that mitigates planning errors caused by stale historical priors and longitudinal motion-stage transitions.

    \item We introduce \textbf{SMM} to suppress stale cross-cycle planning priors and \textbf{MSTS} to learn phase-wise future motion, and fuse their complementary endpoints without additional inference-time computation.
    
\item We introduce \textbf{LT-nuScenes}, a ground-truth-based subset for evaluating planning robustness under longitudinal motion transitions. Our StableDrive achieves \textbf{SOTA} performance on nuScenes, LT-nuScenes, and NAVSIMv1/2, outperforming our reproduced MomAD~\cite{song2025momad} baseline across every reported metric from 1 to 6~s.
\end{itemize}

\section{Related Work}
\label{sec:related}

\subsection{End-to-End Autonomous Driving}
End-to-end autonomous driving has evolved from direct control imitation toward structured planners that unify perception, prediction, mapping, occupancy, and trajectory generation~\cite{bojarski2016end,codevilla2018cil,wu2022tcp,hu2023uniad,jiang2023vad,li2024sparsedrive,jia2025drivetransformer,tits_dong2026_monocular,tits_yin2026_navdrive,tits_kang2026_planning}. Recent advances further improve planning through generative trajectory modeling and constraint-aware refinement~\cite{liao2025diffusiondrive,zhang2025carplanner,zhang2026diffusionforcing,song2025diver,liu2025guideflow,tits_liu2025_intentiondiffusion,tits_lin2025_contingency,tits_chen2026_contingency}, while representative world-model approaches provide future-aware interaction or reasoning signals~\cite{zheng2024genad,zheng2025world4drive,graphworld2026}. Complementary forecasting and prediction--planning research strengthens multimodal intent modeling, uncertainty management, interaction awareness, and scene understanding~\cite{gu2020vectornet,shi2022mtr,nayakanti2023wayformer,zhou2023qcnet,tits_liang2025_interaction,tits_shao2025_uncertainty,tits_yang2026_guide,tits_wang2026_interaction,tits_xiao2025_multivehicle,tits_wang2026_trpvfusion}. Evaluation resources have also expanded beyond open-loop log replay to closed-loop, data-driven simulation, and long-tail end-to-end driving benchmarks~\cite{caesar2021nuplan,dauner2024navsim,jia2024bench2drive,xu2026wode2e}. However, these methods mainly improve trajectory generation, future reasoning, or perception robustness, without explicitly determining whether a self-generated ego-planning state remains reliable across planning cycles. StableDrive addresses this gap by selectively exploiting historical planning states and introducing long-horizon training guidance without enlarging the deployed inference path.

\subsection{Temporal Memory for End-to-End Planning}
Temporal context has increasingly been incorporated into end-to-end planning through trajectory matching, planning-query momentum, historical prediction queries, autoregressive consistency, sparse memory, and coupled future-scene rollouts. Representative methods include MomAD~\cite{song2025momad}, BridgeAD~\cite{zhang2025bridgead}, CarPlanner~\cite{zhang2025carplanner}, Echo~\cite{sun2025echo}, HiP-AD~\cite{tang2025hipad}, Diffusion Forcing Planner~\cite{zhang2026diffusionforcing}, and ProDrive~\cite{fu2026prodrive}. MFPAD~\cite{wu2026mfpad} further foregrounds memory forgetting for long-horizon end-to-end planning. Adjacent T-ITS studies connect interaction-aware prediction or comprehensive uncertainty estimation to downstream planning~\cite{tits_liang2025_interaction,tits_shao2025_uncertainty}, but do not examine the reliability of cached self-generated planning queries across cycles. While these methods demonstrate the value of planning history, the conditional reliability of self-generated planning states under longitudinal motion transitions remains underexplored. This issue is particularly important during braking, stopping, restarting, or changes in car-following behavior, when velocity trends encoded in earlier plans may no longer reflect the current driving context. StableDrive addresses this gap through a causal memory pathway that assigns an input-dependent reliability weight to the immediately preceding self-predicted planning state before integrating it with the current planning query.

\subsection{State-Space Models for Autonomous Driving}
Structured state-space models provide recurrent, input-dependent updates as an alternative to attention over long temporal contexts~\cite{gu2022s4,tits_liu2025_bevmamba}. Mamba~\cite{gu2023mamba} introduces selective input-dependent updates, and Mamba-2~\cite{dao2024mamba2} relates this family to structured state-space duality. Driving applications include Vision Mamba~\cite{zhu2024visionmamba} and VMamba~\cite{liu2024vmamba} for visual representation, MambaOcc~\cite{tian2024mambaocc} for occupancy prediction, MamBEV~\cite{ke2025mambev} for BEV representation, Trajectory Mamba~\cite{huang2025trajectorymamba} and MLSTP~\cite{tits_ren2026_mlstp} for motion forecasting, GMF-Drive~\cite{wang2025gmfdrive} for feature fusion, and DriveMamba~\cite{drivemamba2026} for end-to-end decoding. In most cases the recurrent state represents scene features or generic motion context. StableDrive uses a Mamba~\cite{gu2023mamba} selective state-space operator inside SMM to update a self-predicted planning state, and separately uses a horizon-wise Mamba~\cite{gu2023mamba} block within MSTS only during training. The distinction is therefore functional rather than a backbone substitution.

\section{Preliminary}
\label{sec:prelim}

\noindent\textbf{Momentum Planning.}
Momentum planning reuses the preceding planning state to improve temporal continuity across adjacent cycles. We consider MomAD~\cite{song2025momad} as a representative formulation. Let $\mathbf{Q}^{p}_{t-1}\in\mathbb{R}^{N_q\times D}$ and $\mathbf{S}^{p}_{t-1}\in\mathbb{R}^{N_q}$ denote the planning queries and their scores from the previous cycle, and let $\mathbf{q}^{p}_{t}\in\mathbb{R}^{D}$ denote the current planning query. Historical planning information is encoded and transferred as
\begin{equation}
\begin{aligned}
\mathbf{M}_{t-1}
&=
\operatorname{LSTM}\!\left(
\sigma(\mathbf{S}^{p}_{t-1})\odot
\operatorname{MLP}(\mathbf{Q}^{p}_{t-1})
\right),                                                     \\
\boldsymbol{\alpha}_{t}
&=
\operatorname{softmax}\!\left(
\frac{
(\mathbf{q}^{p}_{t}\mathbf{W}_{Q})
(\mathbf{M}_{t-1}\mathbf{W}_{K})^{\top}
}{
\sqrt{d_a}
}
\right),                                                     \\
\widetilde{\mathbf{q}}^{p}_{t}
&=
\boldsymbol{\alpha}_{t}
(\mathbf{M}_{t-1}\mathbf{W}_{V}),
\end{aligned}
\label{eq:momentum_planning}
\end{equation}
where $\sigma(\cdot)$ denotes the sigmoid function, $\odot$ denotes element-wise multiplication with the score weights broadcast along the feature dimension, and $d_a$ is the attention dimension. The previous candidate scores modulate the historical modes before LSTM encoding, after which the current query aggregates the encoded history through normalized attention.

\noindent\textbf{Limitation of Momentum Planning.}
The previous-cycle scores indicate relative preference among historical candidates rather than whether the cached planning state remains reliable under the current context. Because the attention in Eq.~\eqref{eq:momentum_planning} is normalized over the cached modes, it only redistributes their relative contributions and does not explicitly determine, from the current context, whether the cached history as a whole should be suppressed. When the current motion stage changes, the motion trend encoded in the preceding plan may become stale, propagating prediction errors or outdated motion assumptions into the current plan. This limitation motivates explicit control over the influence of the preceding planning state in the current cycle.

\begin{figure*}[!t]
  \centering
  \includegraphics[width=\textwidth]{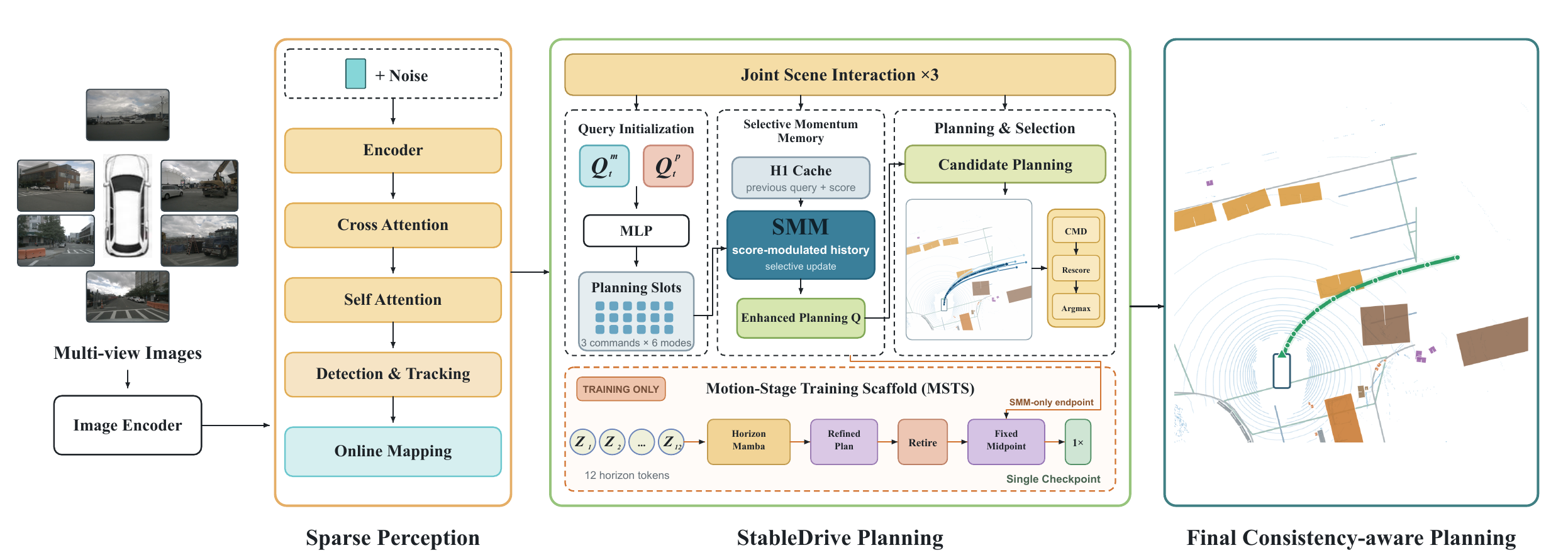}
  \caption{\textbf{Overall framework of StableDrive.} StableDrive combines selective cross-cycle planning memory with a train-and-retire motion-stage scaffold. Multi-view images are encoded into a sparse scene representation, and joint scene interaction generates command-conditioned planning queries. SMM selectively updates the current queries using score-modulated one-cycle history before candidate generation and selection. MSTS provides horizon-wise motion-stage supervision during training and is removed at inference. A fixed parameter midpoint between the SMM-only and MSTS-trained endpoints yields a single deployable checkpoint. All visualizations correspond to the same nuScenes validation sample.}
  \label{fig:framework}
\end{figure*}

\begin{figure}[!t]
  \centering
  \includegraphics[width=0.95\columnwidth]{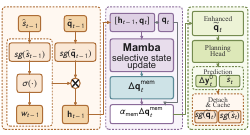}
  \caption{\textbf{Selective Momentum Memory (SMM).} SMM gates the one-cycle cached query with its previous score, integrates the retained history with the current query through a causal Mamba~\cite{gu2023mamba} update, and injects the result as a residual. The refined query drives trajectory prediction, after which it and the updated score are detached and stored for the next cycle.}
  \label{fig:methoddetails}
\end{figure}

\section{Method}
\label{sec:method}

\subsection{Overview}
StableDrive augments the inherited planning pipeline with \textbf{Selective Momentum Memory (SMM)} and a training-only \textbf{Motion-Stage Training Scaffold (MSTS)}. SMM regulates the influence of the preceding self-predicted planning state across planning cycles, while MSTS provides motion-stage-aware training guidance within the prediction horizon. After training, MSTS is removed. A fixed parameter midpoint between the SMM-only and MSTS-trained endpoints then yields a single deployable SMM planner. \Cref{fig:framework} summarizes the overall framework.

StableDrive follows the sparse perception and planning pipeline of MomAD~\cite{song2025momad} and SparseDrive~\cite{li2024sparsedrive}. Multi-view images are encoded into sparse agent and map instances, which are used by joint scene interaction to produce command- and mode-conditioned planning slots. We denote these slots as
\begin{equation}
  \mathbf{Q}_{t}
  \equiv
  \mathbf{Q}_{t}^{p}
  =
  \{\mathbf{q}_{t,c,k}\}_{c=1,k=1}^{C,K}
  \in\mathbb{R}^{C\times K\times D},
\end{equation}
where $C=3$ is the number of command branches, $K=6$ is the number of trajectory modes per command, and $D$ is the query dimension. SMM refines each planning query using the valid one-cycle cache. The planning head then predicts displacement increments $\Delta\mathbf{Y}^{0}_{t}\in\mathbb{R}^{C\times K\times H\times2}$ and candidate scores $\mathbf{S}_{t}\in\mathbb{R}^{C\times K}$, where $H=12$ future points are sampled at 0.5-s intervals. The prescribed command branch is processed by the inherited collision-aware rescoring and argmax selection to obtain the selected plan. Query initialization, the candidate-scoring head, and the final selection procedure remain unchanged from the base planner.

\subsection{Selective Momentum Memory}
In Eq.~\eqref{eq:momentum_planning}, normalized attention redistributes influence among the cached modes but does not explicitly represent rejecting the historical state as a whole. SMM instead separates candidate-level score modulation from context-dependent state integration and uses a planning-specific causal state update to control how strongly the preceding self-prediction affects the current query. The H1 cache in \Cref{fig:framework} denotes the valid one-cycle query--score cache used by this selective update. The complete SMM update and cache flow is summarized in \Cref{fig:methoddetails}.

Let $\widehat{\mathbf{q}}_{t-1,c,k}$ denote the SMM-enhanced planning query immediately before the trajectory regression and scoring heads at cycle $t-1$, and let $\widehat{s}_{t-1,c,k}$ be its corresponding score logit. The pair $(c,k)$ denotes a fixed command--trajectory-anchor slot preserved across adjacent planning cycles; all $CK=18$ slots are cached and detached before use by the next cycle. Thus, the cache is neither a selected final trajectory nor an unordered set of candidate features. For valid history, the score-modulated history token is
\begin{equation}
  \mathbf{h}_{t-1,c,k}
  =
  \sigma(\widehat{s}_{t-1,c,k})
  \operatorname{sg}(\widehat{\mathbf{q}}_{t-1,c,k}),
  \qquad r_t=1,
  \label{eq:smm_history}
\end{equation}
where $\operatorname{sg}(\cdot)$ denotes stop-gradient and $r_t\in\{0,1\}$ indicates whether the cache is the valid preceding frame of the same scene. The score logit provides candidate-level attenuation, while the subsequent input-dependent state transition controls how strongly the preceding state influences the current query.

For every command--trajectory-anchor slot, the modulated history precedes the current query in a causal sequence:
\begin{equation}
  \mathbf{Z}^{\mathrm{mem}}_{t,c,k}
  =
  [\mathbf{h}_{t-1,c,k},\mathbf{q}_{t,c,k}]
  \in\mathbb{R}^{2\times D},
  \qquad r_t=1.
  \label{eq:smm_sequence}
\end{equation}
A Mamba selective state-space operator~\cite{gu2023mamba} processes the sequence:
\begin{equation}
  \mathbf{U}^{\mathrm{mem}}_{t,c,k}
  =
  \Phi_{\mathrm{mem}}
  (\mathbf{Z}^{\mathrm{mem}}_{t,c,k}).
  \label{eq:smm_mamba}
\end{equation}
Its input-dependent transition conditions the terminal state jointly on the preceding self-prediction and the current planning context. SMM then applies a near-identity residual readout with an explicit no-history bypass:
\begin{equation}
\widetilde{\mathbf{q}}_{t,c,k}
=
\begin{cases}
\mathbf{q}_{t,c,k}
+
\alpha_{\mathrm{mem}}
\operatorname{LN}
\left(
\mathbf{U}^{\mathrm{mem}}_{t,c,k}[-1]
\right),
& r_t=1,\\[3pt]
\mathbf{q}_{t,c,k},
& r_t=0,
\end{cases}
\label{eq:smm_readout}
\end{equation}
where $\alpha_{\mathrm{mem}}$ is learned jointly with the planner and initialized to $10^{-3}$. This initialization keeps the initial update close to the original planner, while the $r_t=0$ branch exactly matches the implementation-level identity path.

After the current heads produce their outputs, all slots are written to the next cache as
\begin{equation}
  \widehat{\mathbf{q}}_{t,c,k}
  \leftarrow
  \operatorname{sg}(\widetilde{\mathbf{q}}_{t,c,k}),
  \qquad
  \widehat{s}_{t,c,k}
  \leftarrow
  \operatorname{sg}(s_{t,c,k}).
  \label{eq:smm_cache_write}
\end{equation}
The cache is rank-local and scene-isolated and is reset when history is missing, the scene changes, or temporal ordering is discontinuous. No ground-truth trajectory, future observation, or future feature enters this pathway.

\subsection{Motion-Stage Training Scaffold}
\label{sec:scaffold}
\begin{figure}[!t]
  \centering
  \includegraphics[width=\columnwidth]{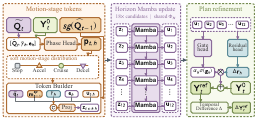}
  \caption{\textbf{Motion-Stage Training Scaffold (MSTS).} The predicted motion-stage distribution conditions candidate-specific horizon tokens. A shared Horizon Mamba~\cite{gu2023mamba} processes each candidate sequence without cross-candidate state exchange, after which gated residual refinement produces the refined trajectory increments. MSTS is training-only, and ground truth is used only for supervision.}
  \label{fig:mstsmechanism}
\end{figure}

SMM regulates planning-state transfer across cycles but does not organize motion evolution over the $H$ future steps. During training, MSTS supplies this within-horizon structure by constructing a motion token at each step from the SMM-enhanced query, the coarse trajectory, and the detached one-cycle query cache. A shared Horizon Mamba models each candidate sequence independently, as illustrated in \Cref{fig:mstsmechanism}. \Cref{alg:msts} gives the forward computation; the supervision and training properties are described below.

\begin{figure}[!t]
\begin{ruledalgorithm}{Motion-Stage Training Scaffold (MSTS)}
\label{alg:msts}
\noindent\textbf{Input:} SMM-enhanced queries $\widetilde{\mathbf Q}_{t}$, coarse trajectory increments $\Delta\mathbf Y^{0}_{t}$, and one-cycle query cache $\widehat{\mathbf Q}_{t-1}$ with validity indicator $r_t$.\\
\textbf{Output:} Refined increments $\Delta\mathbf Y^{\mathrm{ref}}_{t}$ and phase probabilities $\{\mathbf p_{t,h}\}_{h=1}^{H}$.

\begin{enumerate}[leftmargin=1.45em,label=\scriptsize\arabic*,itemsep=2pt,topsep=3pt,parsep=0pt]
\item \textbf{Construct the motion representation:}
\begin{equation*}
\begin{aligned}
\mathbf y^{0}_{t,c,k,h}
&=\sum_{\tau=1}^{h}\Delta\mathbf y^{0}_{t,c,k,\tau}
=(x^{0}_{t,c,k,h},y^{0}_{t,c,k,h}),\\
v^{0}_{t,c,k,h}
&=\Delta y^{0}_{t,c,k,h}/\Delta t,\\
a^{0}_{t,c,k,h}
&=(v^{0}_{t,c,k,h}-v^{0}_{t,c,k,h-1})/\Delta t,
\quad a^{0}_{t,c,k,1}=0,\\
\bm{\gamma}_{t,c,k,h}
&=\operatorname{Concat}[
\mathbf y^{0}_{t,c,k,h},
\Delta\mathbf y^{0}_{t,c,k,h},\\
&\qquad
y^{0}_{t,c,k,h},x^{0}_{t,c,k,h},
v^{0}_{t,c,k,h},a^{0}_{t,c,k,h}]
\in\mathbb R^{8}.
\end{aligned}
\end{equation*}
Here $\Delta t=\SI{0.5}{s}$; $x$ is lateral and positive $y$ is forward.

\item \textbf{Estimate the shared stage context:}
\begin{equation*}
\begin{aligned}
\overline{\widetilde{\mathbf q}}_{t}
&=\frac{1}{CK}\sum_{c,k}\widetilde{\mathbf q}_{t,c,k},
&
\overline{\bm{\gamma}}_{t,h}
&=\frac{1}{CK}\sum_{c,k}\bm{\gamma}_{t,c,k,h},\\
\mathbf p_{t,h}
&=\operatorname{softmax}\!\left(
f_p[\overline{\widetilde{\mathbf q}}_{t},
\overline{\bm{\gamma}}_{t,h},\mathbf e_h]\right),
&
\mathbf u_{t,h}&=W_p\mathbf p_{t,h}.
\end{aligned}
\end{equation*}
The fixed phase order is $\{\text{stationary},\text{accelerating},\text{cruising},\text{decelerating}\}$.

\item \textbf{Construct candidate-specific horizon tokens:}
\begin{equation*}
\begin{aligned}
\mathbf m^{\mathrm{sp}}_{t-1,c,k}
&=r_t\operatorname{sg}(\widehat{\mathbf q}_{t-1,c,k}),\\
\mathbf z_{t,c,k,h}
&=\operatorname{Proj}[
W_q\widetilde{\mathbf q}_{t,c,k},
W_m\mathbf m^{\mathrm{sp}}_{t-1,c,k},
W_\gamma\bm{\gamma}_{t,c,k,h},
\mathbf e_h,\mathbf u_{t,h}].
\end{aligned}
\end{equation*}

\item \textbf{Model horizon evolution:}
\begin{equation*}
\mathbf U^{\mathrm{hor}}_{t,c,k}
=\Phi_{\mathrm{hor}}([
\mathbf z_{t,c,k,1},\ldots,\mathbf z_{t,c,k,H}]),
\qquad C K=18,\ H=12.
\end{equation*}
Parameters are shared across candidates, while no state is exchanged between candidate sequences.

\item \textbf{Refine the trajectory and recover increments:}
\begin{equation*}
\begin{aligned}
\mathbf g_{t,c,k,h}
&=f_g(\mathbf U^{\mathrm{hor}}_{t,c,k,h}),\\
\Delta\mathbf r_{t,c,k,h}
&=f_r(\mathbf U^{\mathrm{hor}}_{t,c,k,h}),\\
\mathbf r^{\mathrm{cum}}_{t,c,k,h}
&=\alpha_h\sigma(\mathbf g_{t,c,k,h})
\odot\Delta\mathbf r_{t,c,k,h},\\
\mathbf y^{\mathrm{ref}}_{t,c,k,h}
&=\mathbf y^{0}_{t,c,k,h}+\mathbf r^{\mathrm{cum}}_{t,c,k,h},\\
\Delta\mathbf y^{\mathrm{ref}}_{t,c,k,h}
&=\mathbf y^{\mathrm{ref}}_{t,c,k,h}
-\mathbf y^{\mathrm{ref}}_{t,c,k,h-1},
\qquad \mathbf y^{\mathrm{ref}}_{t,c,k,0}=\mathbf 0.
\end{aligned}
\end{equation*}
Here $\alpha_h$ is a learned horizon scale.
\end{enumerate}
\end{ruledalgorithm}
\end{figure}

Ground-truth motion stages are constructed only as supervision targets. Let $v^\ast_{t,h}=\Delta y^\ast_{t,h}/\Delta t$ and $a^\ast_{t,h}=(v^\ast_{t,h}-v^\ast_{t,h-1})/\Delta t$, with $a^\ast_{t,1}=0$. A valid horizon is stationary when $|v^\ast_{t,h}|<\SI{0.5}{m/s}$, accelerating when it is non-stationary and $a^\ast_{t,h}>\SI{0.5}{m/s^2}$, decelerating when it is non-stationary and $a^\ast_{t,h}<-\SI{0.5}{m/s^2}$, and cruising otherwise. Invalid horizons are excluded from the corresponding scaffold losses.

The MSTS-assisted training objective is
\begin{equation}
\begin{aligned}
\mathcal{L}_{\mathrm{MSTS}}
&=\mathcal{L}_{\mathrm{base}}
+0.05\mathcal{L}_{\mathrm{phase}}
+0.10\mathcal{L}_{\mathrm{far}}
+0.10\mathcal{L}_{\mathrm{long}},\\
\mathcal{L}_{\mathrm{long}}
&=0.5\mathcal{L}_{y}
+0.3\mathcal{L}_{v_y}
+0.2\mathcal{L}_{a_y}.
\end{aligned}
\label{eq:msts_objective}
\end{equation}
Here $\mathcal{L}_{\mathrm{base}}$ is the original planning objective evaluated on $\Delta\mathbf Y^{\mathrm{ref}}_{t}$, $\mathcal{L}_{\mathrm{phase}}$ is masked cross-entropy, and $\mathcal{L}_{\mathrm{far}}$ is masked L1 loss on cumulative $xy$ positions at horizons 8--12. The terms $\mathcal{L}_{y}$, $\mathcal{L}_{v_y}$, and $\mathcal{L}_{a_y}$ are masked L1 losses over valid future horizons.

The residual head $f_r$ is initialized to zero, so MSTS begins as an identity refinement. The planning loss on the refined trajectory backpropagates through MSTS into the shared planning head and upstream representation. Ground-truth trajectories are used only to construct supervision targets and losses; they never enter the SMM cache, the phase-head inputs, or the horizon tokens. MSTS-specific parameters are removed before inference, as described in \Cref{sec:retirement}.

\subsection{Train--Retire Endpoint Construction}
\label{sec:retirement}
Starting from the same causal SMM initialization $\theta_{\mathrm{init}}$, two branches are trained under a matched optimization budget $\mathcal{B}$:
\begin{align}
  \theta_{\mathrm{SM}}
  &=
  \mathcal{T}_{\mathrm{SMM}}
  (\theta_{\mathrm{init}};\mathcal{B}),\\
  (\theta_{\mathrm{VT}},\phi_{\mathrm{MSTS}})
  &=
  \mathcal{T}_{\mathrm{MSTS}}
  (\theta_{\mathrm{init}},\phi_0;\mathcal{B}),
  \label{eq:endpoint_training}
\end{align}
where $\phi_{\mathrm{MSTS}}$ contains scaffold-specific tensors. The first branch performs SMM-only continuation. The second jointly optimizes the shared planner and MSTS.

Retirement is a checkpoint-construction operation rather than a runtime switch: all scaffold-specific tensors are discarded, while the MSTS-trained shared planner tensors remain $\theta_{\mathrm{VT}}$. The resulting endpoint has the same parameter count, cache lifecycle, input/output interface, and forward graph as the SMM-only endpoint. Formally,
\begin{equation}
  \begin{aligned}
  \mathcal{R}_{\mathrm{shared}}
  (\theta_{\mathrm{VT}},\phi_{\mathrm{MSTS}})
  &=\theta_{\mathrm{VT}},\\
  f_{\mathrm{VT}}(\mathbf{x})
  &=f_{\mathrm{SM}}(\mathbf{x};\theta_{\mathrm{VT}}),
  \end{aligned}
  \label{eq:retired_endpoint}
\end{equation}
where $f_{\mathrm{SM}}$ denotes the common deployed SMM architecture. The effect of MSTS-assisted optimization is retained in $\theta_{\mathrm{VT}}$, while the scaffold itself is absent from inference.

\subsection{Fixed Midpoint Checkpoint Construction}
\label{sec:midpoint}
The preceding training protocol yields two endpoints with the same deployed SMM architecture. The SMM-only endpoint $\theta_{\mathrm{SM}}$ is optimized without the scaffold, whereas $\theta_{\mathrm{VT}}$ retains the effect of MSTS-assisted training after the scaffold has been removed. The remaining step is to obtain one final checkpoint without deploying both endpoints or retaining MSTS at inference. We therefore construct StableDrive by taking the fixed parameter midpoint of the tensors shared by these two endpoints.

Because both endpoints use the same deployed parameterization, their shared tensors can be matched directly by name, shape, and type. They start from the same SMM initialization and are trained under the matched budget $\mathcal{B}$, providing a controlled basis for tensor-wise interpolation. This construction requires neither permutation matching nor output-space ensembling.

Let $\mathcal{S}$ denote the set of name-, shape-, and type-matched floating-point tensors shared by $\theta_{\mathrm{SM}}$ and $\theta_{\mathrm{VT}}$. For $\alpha\in[0,1]$, define the interpolated shared tensors as
\begin{equation}
  \theta_{\alpha}[j]
  =
  (1-\alpha)\theta_{\mathrm{SM}}[j]
  +\alpha\theta_{\mathrm{VT}}[j],
  \qquad \alpha\in[0,1],
  \qquad j\in\mathcal{S}.
  \label{eq:midpoint_construction}
\end{equation}
StableDrive uses $\theta_{\mathrm{SD}}=\theta_{0.5}$. On the interpolation segment, this is the unique point equidistant from the two matched endpoints under the Euclidean norm:
\begin{equation}
  \left\|\theta_{0.5}-\theta_{\mathrm{SM}}\right\|_2
  =
  \left\|\theta_{0.5}-\theta_{\mathrm{VT}}\right\|_2
  =
  \tfrac{1}{2}
  \left\|\theta_{\mathrm{VT}}-\theta_{\mathrm{SM}}\right\|_2.
  \label{eq:midpoint_equidistance}
\end{equation}
Thus, $\alpha=0.5$ gives both training endpoints equal geometric weight and minimizes the maximum Euclidean distance to either endpoint along the segment. This symmetry is a data-independent construction rule rather than a claim of global metric optimality: $\alpha=0.5$ is fixed before final-model evaluation and is neither learned nor selected through a post-hoc metric search. The identical non-floating buffers are copied unchanged, while MSTS-specific tensors, optimizer states, and scheduler states are excluded. The resulting planner is
\begin{equation}
  f_{\mathrm{SD}}(\mathbf{x})
  =
  f_{\mathrm{SM}}(\mathbf{x};\theta_{\mathrm{SD}}).
\end{equation}
The resulting $f_{\mathrm{SD}}$ uses the common deployed SMM computation graph, contains no MSTS-specific tensors, and requires one checkpoint and one forward pass at inference.

\begin{table*}[t]
\centering
\caption{
Planning performance on the nuScenes validation set over the complete 1--6\,s planning horizon.
All methods use camera-only input.
Avg. denotes the mean over the six horizons; for TPC, only the six-horizon average is reported due to space constraints.
R50/R101 denote ResNet-50/101, and DiT denotes Diffusion Transformer.
FPS is provided only as a hardware-specific efficiency reference; GPU subscripts indicate the device used for measurement, and FPS values are not compared across devices.
``--'' denotes unavailable results.
Lower is better for L2, Col. Rate, and TPC, and the best planning results are shown in bold.
}
\label{tab:external}
\begingroup
\scriptsize
\setlength{\tabcolsep}{0.8pt}
\renewcommand{\arraystretch}{1.08}
\begin{tabularx}{\textwidth}{@{}p{2.35cm}>{\centering\arraybackslash}p{0.90cm}*{6}{>{\centering\arraybackslash}X}>{\columncolor{gray!10}\centering\arraybackslash}X*{6}{>{\centering\arraybackslash}X}>{\columncolor{gray!10}\centering\arraybackslash}X>{\columncolor{gray!10}\centering\arraybackslash}p{0.64cm}>{\centering\arraybackslash}p{0.80cm}@{}}
\toprule
Method & Backb. & \multicolumn{7}{c}{L2 (m) $\downarrow$} & \multicolumn{7}{c}{Col. Rate (\%) $\downarrow$} & \cellcolor{white}TPC$\downarrow$ & FPS$\uparrow$\\
\cmidrule(lr){3-9}\cmidrule(lr){10-16}\cmidrule(lr){17-17}\cmidrule(l){18-18}
& & 1s & 2s & 3s & 4s & 5s & 6s & \cellcolor{white}Avg. & 1s & 2s & 3s & 4s & 5s & 6s & \cellcolor{white}Avg. & \cellcolor{white}Avg. & \\
\midrule
UniAD~\cite{hu2023uniad} & R101 & 0.47 & 0.91 & 1.35 & 1.91 & 2.47 & 3.07 & 1.70 & 0.25 & 0.36 & 0.61 & 0.99 & 1.64 & 2.51 & 1.06 & 1.58 & \resizebox{0.72cm}{!}{1.8$_{\rm A100}$}\\
VAD~\cite{jiang2023vad} & R50 & 0.41 & 0.70 & 1.05 & 1.82 & 2.23 & 3.01 & 1.54 & 0.03 & 0.19 & 0.43 & 0.89 & 1.71 & 2.41 & 0.94 & 1.23 & 4.5$_{\rm 3090}$\\
SparseDrive~\cite{li2024sparsedrive} & R50 & 0.43 & 0.87 & 1.23 & 1.75 & 2.32 & 2.95 & 1.59 & 0.19 & 0.31 & 0.56 & 0.87 & 1.54 & 2.33 & 0.97 & 1.46 & 9.0$_{\rm 4090}$\\
DiffusionDrive~\cite{liao2025diffusiondrive} & R50 & 0.35 & 0.81 & 0.99 & 1.65 & 1.94 & 2.40 & 1.36 & 0.20 & 0.32 & 0.45 & 0.82 & 1.51 & 2.23 & 0.92 & 1.31 & 8.2$_{\rm 4090}$\\
DIVER~\cite{song2025diver} & R50 & 0.38 & 0.75 & 1.10 & 1.53 & 1.98 & 2.49 & 1.37 & 0.13 & 0.31 & 0.44 & 0.80 & 1.41 & 2.11 & 0.87 & 1.35 & --\\
GuideFlow~\cite{liu2025guideflow} & R50 & 0.42 & 0.83 & 1.21 & 1.73 & 2.05 & 2.63 & 1.48 & 0.12 & 0.22 & 0.42 & 0.79 & 1.44 & 2.15 & 0.86 & 1.34 & 3.6$_{\rm 4090}$\\
LAW~\cite{li2025law} & Swin-T & 0.40 & 0.87 & 1.16 & 1.71 & 2.03 & 2.61 & 1.46 & 0.19 & 0.33 & 0.57 & 0.86 & 1.51 & 2.31 & 0.96 & -- & --\\
Epona~\cite{zhang2025epona} & DiT & 0.39 & 0.91 & 1.17 & 1.73 & 2.02 & 2.75 & 1.50 & 0.14 & 0.18 & 0.45 & 0.74 & 1.48 & 2.23 & 0.87 & -- & --\\
World4Drive~\cite{zheng2025world4drive} & R50 & 0.42 & 0.92 & 1.21 & 1.75 & 2.06 & 2.79 & 1.53 & 0.16 & 0.20 & 0.47 & 0.76 & 1.50 & 2.14 & 0.87 & -- & --\\
MomAD~\cite{song2025momad} & R50 & 0.41 & 0.85 & 1.13 & 1.67 & 1.98 & 2.45 & 1.41 & 0.17 & 0.30 & 0.54 & 0.83 & 1.43 & 2.13 & 0.90 & 1.30 & 7.8$_{\rm 4090}$\\
\rowcolor{gray!15}
\textbf{StableDrive (Ours)} & R50 & \textbf{0.28} & \textbf{0.55} & \textbf{0.89} & \textbf{1.32} & \textbf{1.80} & \textbf{2.35} & \textbf{1.20} & \textbf{0.01} & \textbf{0.09} & \textbf{0.30} & \textbf{0.65} & \textbf{1.16} & \textbf{1.78} & \textbf{0.66} & \textbf{0.85} & 5.2$_{\rm 4090}$\\
\bottomrule
\end{tabularx}
\endgroup
\end{table*}

\begin{table*}[t]
\centering
\caption{Planning on LT-nuScenes over 1--6\,s. MomAD$^{\dagger}$~\cite{song2025momad} is our local matched six-second reproduction. Avg. is the six-horizon mean; TPC is shown only as Avg. Red subscripts are StableDrive minus MomAD$^{\dagger}$; negative values indicate improvement, and Col. Rate differences are percentage points. Lower is better; bold marks the best displayed values, including rounded ties.}
\label{tab:lt_nuscenes}
\begingroup
\scriptsize
\setlength{\tabcolsep}{1.0pt}
\renewcommand{\arraystretch}{1.08}
\newcommand{\tabledelta}[2]{\resizebox{0.80cm}{!}{\(\mathbf{#1}_{\textcolor{red}{\scriptscriptstyle #2}}\)}}
\begin{tabularx}{\textwidth}{@{}>{\raggedright\arraybackslash}p{2.65cm}*{14}{>{\centering\arraybackslash}X}>{\centering\arraybackslash}p{1.35cm}@{}}
\toprule
Method & \multicolumn{7}{c}{L2 (m) $\downarrow$} & \multicolumn{7}{c}{Col. Rate (\%) $\downarrow$} & \cellcolor{white}TPC (m) $\downarrow$\\
\cmidrule(lr){2-8}\cmidrule(lr){9-15}\cmidrule(l){16-16}
& 1~s & 2~s & 3~s & 4~s & 5~s & 6~s & \cellcolor{white}Avg. & 1~s & 2~s & 3~s & 4~s & 5~s & 6~s & \cellcolor{white}Avg. & \cellcolor{white}Avg.\\
\midrule
MomAD$^\dagger$ & \textbf{0.30} & \textbf{0.62} & 1.04 & 1.53 & 2.07 & 2.65 & 1.37 & 0.53 & 1.59 & 2.56 & 3.64 & 4.60 & 5.56 & 3.08 & 0.85\\
\rowcolor{gray!15}
\textbf{StableDrive (Ours)} & \tabledelta{0.30}{0.00} & \tabledelta{0.62}{0.00} & \tabledelta{1.03}{-0.01} & \tabledelta{1.50}{-0.03} & \tabledelta{1.98}{-0.09} & \tabledelta{2.48}{-0.17} & \tabledelta{1.32}{-0.05} & \tabledelta{0.00}{-0.53} & \tabledelta{0.26}{-1.33} & \tabledelta{0.44}{-2.12} & \tabledelta{1.26}{-2.38} & \tabledelta{2.75}{-1.85} & \tabledelta{4.23}{-1.33} & \tabledelta{1.49}{-1.59} & \tabledelta{0.79}{-0.06}\\
\bottomrule
\end{tabularx}
\endgroup
\end{table*}

\begin{table}[t]
\centering
  \caption{\textcolor{black}{Comparison on planning-oriented \textbf{NAVSIMv1} navtest split with $\operatorname{\textbf{Closed-Loop}}$ metrics. `mmt' refers multi-modal trajectory variant of $\operatorname{TransFuser}$ and $^*$ denotes the re-implementation. }}
\renewcommand\arraystretch{0.7}
  \tabcolsep=1.6mm 
  \resizebox{\linewidth}{!}{
    \begin{tabular}{l c c c c c c}
    \toprule
    \multicolumn{1}{l}{Method}&  NC$\uparrow$& DAC$\uparrow$& TTC$\uparrow$& Conf.$\uparrow$& EP$\uparrow$& PDMS$\uparrow$ \\
            \midrule
            VADv2~\cite{chen2024vadv2} &  97.2 & 89.1 & 91.6 & 100 & 76.0 & 80.9 \\
            TransFuser~\cite{TransFuser} &  97.7 & 92.8 & 92.8 & 100 & 79.2 & 84.0 \\
            ${\operatorname{TransFuser}_{\operatorname{mmt}}}^{*}$ \cite{TransFuser}&96.2& 95.4& 90.7& 100& 80.7& 85.1 \\ 
            UniAD~\cite{hu2023uniad} &  97.8 & 91.9 & 92.9 & 100 & 78.8 & 83.4 \\
            PARA-Drive~\cite{paradrive} &  97.9 & 92.4 & 93.0 & 99.8 & 79.3 & 84.0 \\
            DRAMA~\cite{yuan2024drama} &  98.0 & 93.1 & 94.8 & 100 & 80.1 & 85.5 \\
            GoalFlow~\cite{xing2025goalflow}  & 98.3 & 93.8 & 94.3 & 100 & 79.8 & 85.7 \\
            Hydra-MDP~\cite{li2024hydra} &  98.3 & 96.0 & 94.6 & 100 & 78.7 & 86.5 \\
            FUMP~\cite{liu2025fump} &  98.1 & 96.2 & 94.2 & 100 & 82.0 & 87.8 \\
            DiffusionDrive~\cite{liao2025diffusiondrive} & 98.2 & 96.2 & 94.7 & 100 & 82.2 & 88.1 \\
            DIVER~\cite{song2025diver} &  98.5 & 96.5 & 94.9 & 100 & 82.6 & 88.3 \\
            DriveSuprim~\cite{yao2025drivesuprim} &  97.8 & 97.3 & 93.6 & 100 & 86.7 & 89.9 \\
            GoalFlow~\cite{xing2025goalflow}  &  98.4 & 98.3 & 94.6 & 100 & 85.0 & 90.3 \\
            ReCogDrive-IL\cite{li2025recogdrive}&98.1& 94.7 &94.2& 100& 80.9 &86.5\\
            \rowcolor{gray!15}
\textbf{StableDrive (Ours)} &   99.1& 97.0&95.2& 100 &82.9 &90.4\\ 
            \bottomrule
    \end{tabular}}
\label{tab_NAVSIMv1}
\end{table}

\begin{table}[t]
\centering
    \caption{Comparison with SOTA methods on the \textbf{NAVSIMv2 navtest} split~\cite{navsimv2}. EPDMS$^{*}$ denotes results computed with the original NAVSIMv2 evaluation code before the human-behavior filtering fix, while EPDMS denotes the corrected official implementation.}
\renewcommand\arraystretch{0.7}
  \tabcolsep=0.8mm 
  \resizebox{\linewidth}{!}{
\begin{tabular}{l c ccccc cccc}
\toprule
\multicolumn{1}{l}{Method}&  NC$\uparrow$& DAC$\uparrow$& DDC$\uparrow$& TL$\uparrow$& EP$\uparrow$& TTC$\uparrow$& LK$\uparrow$& HC$\uparrow$& EC$\uparrow$& EPDMS$\uparrow$ \\
        \midrule

        TransFuser~\cite{TransFuser} & 96.9 & 89.9 & 97.8 & 99.7 & 87.1 & 95.4 & 92.7 & 98.3 & 87.2 & 76.7 \\
        DiffusionDrive~\cite{liao2025diffusiondrive} & 98.2 & 95.9 & 99.4 & 99.8 & 87.5 & 97.3 & 96.8 & 98.3 & 87.7  &84.5 \\
        Hydra-MDP++~\cite{li2025hydraplus} & 97.2 & 97.5 & 99.4 & 99.6 & 83.1 & 96.5 & 94.4 & 98.2 & 70.9 & 81.4  \\
        DriveSuprim~\cite{yao2025drivesuprim} & 97.5 & 96.5 & 99.4 & 99.6 & 88.4 & 96.6 & 95.5 & 98.3 & 77.0 & 83.1  \\
        DiffusionDriveV2~\cite{zou2025diffusiondrivev2} & 97.7 & 96.6 & 99.2 & 99.8 & 88.9 & 97.2 & 96.0 & 97.8 & 91.0 &  87.5 \\
        DriveWorld-VLA~\cite{liu2026driveworld} & 98.6 & 99.1 & 99.6 & 99.8 & 87.4 & 97.9 & 97.0 & 97.8 & 78.6 & 86.8 \\
        DriveVLA-W0~\cite{li2025drivevla} & 98.5 & 99.1 & 98.0 & 99.7 & 86.4 & 98.1 & 93.2 & 97.9 & 58.9 &  86.1 \\
        Recogdrive~\cite{li2025recogdrive} & 98.3 & 95.2 & 98.3 & 99.8 & 87.1 & 97.5 & 96.6 & 99.5 & 86.5 &  83.6 \\
        Latent-WAM~\cite{wang2026latentwam} & 98.1 & 97.3 & 99.6 & 99.8 & 87.7 & 97.3 & 97.6 & 98.1 & 87.3 &  89.3 \\
        \rowcolor{gray!15} StableDrive (Ours) & 98.7 &99.1 &99.4& 99.0& 86.6 &97.5& 96.4& 98.2 &87.6&  90.0 \\
        \bottomrule
\end{tabular}}
\label{tab_navsimv2_navtest}
\end{table}

\begin{table}[t]
\centering
\caption{Comparison with SOTA methods on the \textbf{NAVSIMv2 navhard}
split~\cite{navsimv2}.}
\label{tab_navsimv2_navhard}

\renewcommand{\arraystretch}{0.9}
\setlength{\tabcolsep}{0.3mm}

\resizebox{\linewidth}{!}{%
\begin{tabular}{lc*{10}{c}}
\toprule
Method
& Stage
& NC$\uparrow$
& DAC$\uparrow$
& DDC$\uparrow$
& TL$\uparrow$
& EP$\uparrow$
& TTC$\uparrow$
& LK$\uparrow$
& HC$\uparrow$
& EC$\uparrow$
& EPDMS$\uparrow$ \\
\midrule

\multirow{2}{*}{TransFuser~\cite{TransFuser}}
& Stage 1
& 96.2 & 79.5 & 99.1 & 99.5 & 84.1
& 95.1 & 94.2 & 97.5 & 79.1
& \multirow{2}{*}{23.1} \\
& Stage 2
& 77.7 & 70.2 & 84.2 & 98.0 & 85.1
& 75.6 & 45.4 & 95.7 & 75.9
& \\
\cmidrule(lr){1-12}

\multirow{2}{*}{DiffusionDrive~\cite{liao2025diffusiondrive}}
& Stage 1
& 96.0 & 79.7 & 97.4 & 99.5 & 81.3
& 93.1 & 90.8 & 96.8 & 73.8
& \multirow{2}{*}{24.2} \\
& Stage 2
& 82.1 & 72.2 & 88.5 & 98.7 & 85.1
& 78.8 & 49.2 & 89.3 & 71.2
& \\
\cmidrule(lr){1-12}

\multirow{2}{*}{GuideFlow~\cite{liu2025guideflow}}
& Stage 1
& 96.6 & 80.5 & 96.3 & 99.3 & 82.3
& 94.9 & 91.5 & 97.7 & 67.8
& \multirow{2}{*}{27.1} \\
& Stage 2
& 87.3 & 76.7 & 88.8 & 99.2 & 84.3
& 85.1 & 49.7 & 93.1 & 44.5
& \\
\cmidrule(lr){1-12}

\multirow{2}{*}{RAP-DINO~\cite{feng2026rap}}
& Stage 1
& 97.1 & 94.4 & 98.8 & 99.8 & 83.9
& 96.9 & 94.7 & 96.4 & 66.2
& \multirow{2}{*}{36.9} \\
& Stage 2
& 83.2 & 83.9 & 87.4 & 98.0 & 86.9
& 80.4 & 52.3 & 95.2 & 52.4
& \\
\cmidrule(lr){1-12}

\multirow{2}{*}{MindDrive~\cite{sun2025minddrive}}
& Stage 1
& 96.1 & 86.0 & 98.8 & 99.3 & 83.3
& 95.6 & 94.4 & 97.6 & 74.7
& \multirow{2}{*}{30.9} \\
& Stage 2
& 82.6 & 79.1 & 86.4 & 98.0 & 85.3
& 79.4 & 49.2 & 96.5 & 71.0
& \\
\cmidrule(lr){1-12}

\multirow{2}{*}{World4Drive~\cite{zheng2025world4drive}}
& Stage 1
& 97.3 & 89.1 & 97.6 & 99.7 & 60.5
& 96.8 & 87.7 & 93.1 & 60.0
& \multirow{2}{*}{34.9} \\
& Stage 2
& 91.4 & 82.0 & 91.0 & 98.5 & 53.1
& 90.6 & 52.3 & 93.3 & 62.8
& \\
\cmidrule(lr){1-12}

\rowcolor{gray!15}
& Stage 1
& 98.3 & 97.5 & 97.3 & 98.4 & 84.1
& 97.7 & 97.0 & 95.9 & 65.2
& \\

\rowcolor{gray!15}
\multirow{-2}{*}{StableDrive (Ours)}
& Stage 2
& 88.4 & 86.1 & 93.1 & 98.1 & 83.0
& 87.6 & 56.9 & 92.1 & 44.3
& \multirow{-2}{*}{\textbf{42.6}} \\
\bottomrule
\end{tabular}%
}
\end{table}

\begin{table*}[!t]
\centering
\begin{minipage}[t]{0.485\textwidth}
\refstepcounter{table}
\label{tab:component_ablation}
{\footnotesize
\centering
{\normalfont\footnotesize TABLE~\thetable}\par
\vspace{1pt}
\noindent\parbox[t][3.45\baselineskip][t]{\linewidth}{\centering\normalfont\footnotesize\scshape Component ablation on the nuScenes validation set over 1--6\,s under matched settings. Lower is better; best values are bold.}\par}
\vspace{0.15em}
\begingroup
\scriptsize
\setlength{\tabcolsep}{0.8pt}
\renewcommand{\arraystretch}{1.04}
\begin{tabularx}{\linewidth}{@{}>{\raggedright\arraybackslash}p{3.15cm}*{6}{>{\centering\arraybackslash}X}@{}}
\toprule
Method & 1~s & 2~s & 3~s & 4~s & 5~s & 6~s\\
\midrule
\multicolumn{7}{@{}l}{\textit{L2 (m)} $\downarrow$}\\[-2pt]
MomAD$^\dagger$ & 0.29 & 0.56 & 0.93 & 1.38 & 1.91 & 2.50\\
\smep & 0.30 & 0.59 & 0.97 & 1.44 & 1.97 & 2.58\\
\vtep & 0.29 & 0.57 & 0.93 & 1.36 & 1.86 & 2.42\\
\rowcolor{gray!15}
\textbf{StableDrive (Ours)} & \textbf{0.28} & \textbf{0.55} & \textbf{0.89} & \textbf{1.32} & \textbf{1.80} & \textbf{2.35}\\
\midrule
\multicolumn{7}{@{}l}{\textit{Col. Rate (\%)} $\downarrow$}\\[-2pt]
MomAD$^\dagger$ & 0.04 & 0.20 & 0.47 & 0.89 & 1.43 & 2.05\\
\smep & 0.02 & 0.11 & 0.33 & 0.68 & 1.21 & 1.83\\
\vtep & 0.12 & 0.22 & 0.43 & 0.78 & 1.31 & 1.96\\
\rowcolor{gray!15}
\textbf{StableDrive (Ours)} & \textbf{0.01} & \textbf{0.09} & \textbf{0.30} & \textbf{0.65} & \textbf{1.16} & \textbf{1.78}\\
\midrule
\multicolumn{7}{@{}l}{\textit{TPC (m)} $\downarrow$}\\[-2pt]
MomAD$^\dagger$ & 0.30 & 0.54 & 0.79 & 1.05 & 1.34 & 1.64\\
\smep & 0.29 & 0.50 & 0.73 & 0.98 & 1.23 & 1.49\\
\vtep & 0.29 & 0.50 & 0.73 & 0.97 & 1.22 & 1.48\\
\rowcolor{gray!15}
\textbf{StableDrive (Ours)} & \textbf{0.28} & \textbf{0.49} & \textbf{0.72} & \textbf{0.95} & \textbf{1.20} & \textbf{1.45}\\
\bottomrule
\end{tabularx}
\endgroup
\end{minipage}
\hfill
\begin{minipage}[t]{0.485\textwidth}
\refstepcounter{table}
\label{tab:lt_component_ablation}
{\footnotesize
\centering
{\normalfont\footnotesize TABLE~\thetable}\par
\vspace{1pt}
\noindent\parbox[t][3.45\baselineskip][t]{\linewidth}{\centering\normalfont\footnotesize\scshape Component ablation on LT-nuScenes over 1--6\,s under the same settings. TPC uses 176 frames with valid preceding cycles. Lower is better; best values, including rounded ties, are bold.}\par}
\vspace{0.15em}
\begingroup
\scriptsize
\setlength{\tabcolsep}{0.8pt}
\renewcommand{\arraystretch}{1.04}
\begin{tabularx}{\linewidth}{@{}>{\raggedright\arraybackslash}p{3.15cm}*{6}{>{\centering\arraybackslash}X}@{}}
\toprule
Method & 1~s & 2~s & 3~s & 4~s & 5~s & 6~s\\
\midrule
\multicolumn{7}{@{}l}{\textit{L2 (m)} $\downarrow$}\\[-2pt]
MomAD$^\dagger$ & \textbf{0.30} & \textbf{0.62} & 1.04 & 1.53 & 2.07 & 2.65\\
\smep & \textbf{0.30} & 0.64 & 1.08 & 1.57 & 2.09 & 2.63\\
\vtep & 0.31 & 0.63 & 1.04 & 1.52 & 2.02 & 2.54\\
\rowcolor{gray!15}
\textbf{StableDrive (Ours)} & \textbf{0.30} & \textbf{0.62} & \textbf{1.03} & \textbf{1.50} & \textbf{1.98} & \textbf{2.48}\\
\midrule
\multicolumn{7}{@{}l}{\textit{Col. Rate (\%)} $\downarrow$}\\[-2pt]
MomAD$^\dagger$ & 0.53 & 1.59 & 2.56 & 3.64 & 4.60 & 5.56\\
\smep & \textbf{0.00} & 0.40 & 0.71 & 1.72 & 3.70 & 5.20\\
\vtep & \textbf{0.00} & 0.40 & 1.15 & 2.12 & 3.70 & 5.29\\
\rowcolor{gray!15}
\textbf{StableDrive (Ours)} & \textbf{0.00} & \textbf{0.26} & \textbf{0.44} & \textbf{1.26} & \textbf{2.75} & \textbf{4.23}\\
\midrule
\multicolumn{7}{@{}l}{\textit{TPC (m)} $\downarrow$}\\[-2pt]
MomAD$^\dagger$ & 0.29 & 0.51 & 0.74 & 0.97 & 1.19 & 1.41\\
\smep & 0.29 & 0.50 & 0.72 & 0.93 & 1.13 & 1.32\\
\vtep & \textbf{0.28} & \textbf{0.48} & \textbf{0.69} & \textbf{0.88} & \textbf{1.07} & \textbf{1.24}\\
\rowcolor{gray!15}
\textbf{StableDrive (Ours)} & 0.29 & 0.50 & 0.71 & 0.91 & 1.09 & 1.25\\
\bottomrule
\end{tabularx}
\endgroup
\end{minipage}
\end{table*}

\section{Experiments}
\label{sec:experiments}

\subsection{Dataset}
\noindent\textbf{nuScenes.}
The nuScenes dataset~\cite{caesar2020nuscenes} contains 1,000 scenes with the standard split of 700 training, 150 validation, and 150 test scenes. We evaluate StableDrive on the complete validation split. Under the MomAD~\cite{song2025momad} protocol, the planner predicts 12 future waypoints at \SI{0.5}{s} intervals over a 6-second horizon. The ground-truth future trajectory is used only by the evaluator to determine the prescribed command branch and is never provided to the planner as input.

\noindent\textbf{LT-nuScenes (Ours).}
To evaluate planning under longitudinal motion transitions, we construct \textbf{Longitudinal-Transition nuScenes (LT-nuScenes)}, a curated evaluation subset of the nuScenes validation set. The subset is constructed solely from ground-truth ego kinematics, without using model predictions. Motion states are identified at \SI{0.5}{s} intervals: stationary states satisfy $|v_{\mathrm{lon}}|<\SI{0.5}{m/s}$, while acceleration and deceleration satisfy $a_{\mathrm{lon}}>\SI{0.5}{m/s^2}$ and $a_{\mathrm{lon}}<-\SI{0.5}{m/s^2}$, respectively. A sustained event must last for at least \SI{1.0}{s}. Eligible targets have complete 6-second ground truth and contain stopping, restarting, sustained acceleration, sustained deceleration, or a reversal between acceleration and deceleration. Targets are excluded as turning when the absolute 6-second heading change is at least $15^{\circ}$ and the displacement is at least \SI{2}{m}, or as lane changes when the absolute lateral displacement is at least \SI{1.5}{m} and the forward displacement is at least \SI{5}{m}. A scene is retained only if its valid targets jointly cover stopping, restarting, sustained acceleration, and sustained deceleration, and collectively include transition onsets in the early (0.5--2.0~s), middle (2.5--3.5~s), and far (4.0--6.0~s) horizons. The resulting subset contains 16 complete scenes and 642 inference frames, of which 189 are scored longitudinal-transition targets. The remaining frames are retained during sequential inference to preserve the cross-cycle context required by SMM and TPC.

\noindent\textbf{NAVSIM v1 and v2.} We conduct additional planning-oriented evaluation on NAVSIM v1 and NAVSIM v2 using the official OpenScene-based splits. The filtered \texttt{navtrain} split contains 103,288 training samples from 1,192 logs, while \texttt{navtest} contains 12,146 evaluation samples from 136 logs. Both splits are sampled at 2~Hz. For the more challenging NAVSIM v2 evaluation, we additionally use \texttt{navhard\_two\_stage}, which contains 225 paired two-stage scenario groups from 76 logs. These groups comprise 450 first-stage scene tokens and 5,462 perturbed follow-up tokens, giving 5,912 evaluated tokens in total. The standard \texttt{navtest} split measures performance on the nominal test distribution, whereas \texttt{navhard\_two\_stage} evaluates recovery and rule compliance after the ego vehicle reaches an off-nominal state induced by the first-stage rollout.

\subsection{Evaluation Metrics}
\label{sec:metrics}
\noindent\textbf{nuScenes and LT-nuScenes.} Following MomAD~\cite{song2025momad}, we report L2 trajectory error, collision rate (Col. Rate), and trajectory prediction consistency error (TPC) from 1 to 6~s. L2 measures the displacement between the predicted and ground-truth trajectories, Col. Rate measures the proportion of predicted trajectories that overlap the future occupancy of surrounding agents, and TPC measures the deviation between temporally aligned plans generated in consecutive planning cycles. Lower values indicate better performance for all three metrics. On LT-nuScenes, L2 and Col. Rate are computed on the full scored subset, whereas TPC is computed on the 176 targets with a valid preceding cycle in the same scene.  \par\noindent\textbf{NAVSIM v1.} We report No-at-fault Collision (NC), Drivable Area Compliance (DAC), Time to Collision (TTC), Comfort (Conf.), Ego Progress (EP), and the aggregate Predictive Driver Model Score (PDMS). PDMS treats NC and DAC as multiplicative safety terms and combines TTC, EP, and Conf. using weights of 5, 5, and 2, respectively.  \par\noindent\textbf{NAVSIM v2.} NAVSIM v2 extends this formulation to EPDMS by adding Driving Direction Compliance (DDC) and Traffic-Light Compliance (TL) as multiplicative terms, together with Lane Keeping (LK), History Comfort (HC), and Extended Comfort (EC) as additional weighted terms. The corrected EPDMS implementation further suppresses false-positive penalties when the corresponding human trajectory exhibits the same violation. For \texttt{navhard\_two\_stage}, the first- and second-stage subscores are reported separately. Follow-up scenes in the second stage are weighted according to their proximity to the terminal state produced in the first stage, and the two stages are combined using the official two-stage aggregation. Higher values indicate better performance for all NAVSIM metrics. Results obtained using the original NAVSIM v2 evaluation code are marked as EPDMS*, while EPDMS denotes results computed with the corrected official implementation.

\subsection{Implementation Details}
StableDrive follows the MomAD~\cite{song2025momad} and SparseDrive~\cite{li2024sparsedrive} implementation family, using a ResNet-50~\cite{he2016resnet} image backbone and $704\times256$ multi-view inputs. Training is performed on four GPUs with six samples per GPU, giving a total batch size of 24. We use AdamW~\cite{loshchilov2019adamw} with an initial learning rate of $1.5\times10^{-4}$, a weight decay of $10^{-3}$, 500 linear warm-up updates, cosine annealing~\cite{loshchilov2017sgdr}, and gradient clipping with a maximum $\ell_2$ norm of 25; the image-backbone learning rate is scaled by 0.1. Both endpoint branches start from the same SMM checkpoint and are optimized for 11,720 updates under matched settings. MSTS is removed after training, and the final StableDrive checkpoint is constructed as the fixed arithmetic midpoint ($\alpha=0.5$) of the shared parameters of the two architecture-aligned endpoints. The resulting model uses one checkpoint and one forward pass, without model ensembling or additional inference-time computation. MomAD$^\dagger$ denotes our local six-second reproduction, retrained from the official SparseDrive Stage-2 initialization and evaluated under the same protocol. For the NAVSIM experiments, we retain the StableDrive architecture and deployment procedure described above. The SMM-only and MSTS-assisted endpoints are retrained from the same NAVSIM initialization on \texttt{navtrain} under matched optimization budgets, while \texttt{navtest} and \texttt{navhard\_two\_stage} are reserved exclusively for evaluation. NAVSIM v1 results are computed using the official v1.1 PDMS pipeline. NAVSIM v2 \texttt{navtest} and \texttt{navhard\_two\_stage} results are computed using the corrected v2.2 EPDMS implementation; the latter additionally loads the released synthetic follow-up scenes and applies the official reactive-traffic and two-stage aggregation procedure. After training, MSTS is removed and the two architecture-aligned endpoints are merged using the fixed midpoint with $\alpha=0.5$. Each reported result therefore uses a single checkpoint and one forward pass, without model ensembling, test-time trajectory optimization, or evaluation-set adaptation.

\subsection{Main Results}

\noindent\textbf{nuScenes.}
As shown in \Cref{tab:external}, StableDrive achieves the lowest L2 error and collision rate at every horizon from 1 to 6\,s while obtaining the lowest average TPC. Its average L2, Col. Rate, and TPC are 1.20\,m, 0.66\%, and 0.85\,m, respectively, corresponding to reductions of 11.8\%, 23.3\%, and 30.9\% over the strongest previously reported averages. The consistent gains across prediction horizons show that StableDrive improves long-horizon accuracy, safety, and cross-cycle stability without sacrificing short-horizon performance.  \par\noindent\textbf{LT-nuScenes.} As shown in \Cref{tab:lt_nuscenes}, StableDrive matches or reduces the L2 error of MomAD$^\dagger$ at every horizon from 1 to 6\,s and consistently achieves a lower collision rate. Averaged over the six prediction horizons, StableDrive reduces L2 error from 1.37 to 1.32\,m and Col. Rate from 3.08\% to 1.49\%, corresponding to relative reductions of approximately 3.65\% and 51.62\%, respectively. The average TPC is also reduced from 0.85 to 0.79\,m, yielding a 7.06\% improvement. These results demonstrate the effectiveness of StableDrive under longitudinal motion transitions.

\noindent\textbf{NAVSIM v1 and v2.}
Tables~III--V report results under three settings: NAVSIM v1 \texttt{navtest}, NAVSIM v2 \texttt{navtest}, and NAVSIM v2 \texttt{navhard\_two\_stage}. StableDrive achieves the best aggregate score in all three settings, obtaining 90.4 PDMS, 90.0 EPDMS, and 42.6 EPDMS, respectively. On NAVSIM v1 \texttt{navtest}, it reaches 99.1 NC and 95.2 TTC, showing that the PDMS gain is supported by collision avoidance and forward safety. On NAVSIM v2 \texttt{navtest}, the corresponding 98.7 NC, 99.1 DAC, and 99.4 DDC results indicate consistently strong safety and compliance rather than reliance on a single secondary metric. The advantage is largest on \texttt{navhard\_two\_stage}, where StableDrive exceeds the previous best of 36.9 by 5.7 points. Its first-stage NC and DAC remain high at 98.3 and 97.5, while its second-stage results retain 86.1 DAC and 93.1 DDC and achieve 56.9 LK, the highest value among the compared methods. These results demonstrate robust planning when perturbed follow-up scenes depart from the nominal recorded state. Together with the nuScenes and LT-nuScenes results, the NAVSIM gains show that StableDrive improves long-horizon planning across trajectory-accuracy, temporal-consistency, safety, and rule-compliance criteria.

\subsection{Ablation Study}

\noindent\textbf{Component Contributions.}
\Cref{tab:component_ablation,tab:lt_component_ablation} compare SMM, MSTS-assisted training, and fixed-midpoint construction. On nuScenes, the SMM endpoint reduces Col. Rate and TPC but slightly increases L2, whereas the retired MSTS endpoint improves long-horizon L2 and TPC but is less effective in collision avoidance. StableDrive combines these strengths and ranks first in all 18 reported entries. Relative to MomAD$^\dagger$, it reduces six-horizon-average L2, Col. Rate, and TPC by 5.0\%, 21.5\%, and 10.1\%, respectively. On LT-nuScenes, the retired MSTS endpoint yields the lowest TPC; StableDrive remains within 0.01\,m at 6\,s while improving L2 and Col. Rate.

\begin{figure}[!htbp]
\centering
\includegraphics[width=\columnwidth]{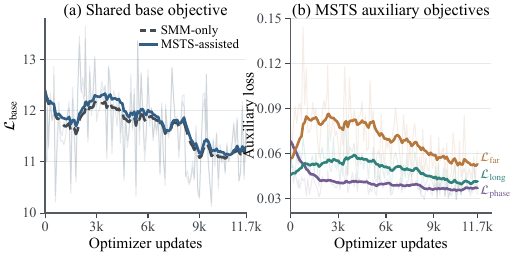}
\caption{\textbf{Training Dynamics under Matched Optimization Budgets.} Both branches start from the same SMM checkpoint and receive 11,720 optimizer updates. \textbf{(a)} The shared base objective $\mathcal{L}_{\mathrm{base}}$ remains comparable and bounded for SMM-only and MSTS-assisted training. \textbf{(b)} The unweighted MSTS auxiliary objectives remain bounded and settle to lower late-training levels, with $\mathcal{L}_{\mathrm{far}}$ exhibiting larger intermediate fluctuations. Faint traces show raw rank-averaged audit values, while the emphasized curves show step-aware EMA trends. MSTS is retired before fixed-midpoint StableDrive construction.}
\label{fig:training_dynamics}
\end{figure}

\begin{figure}[!htbp]
\centering
\includegraphics[width=\columnwidth]{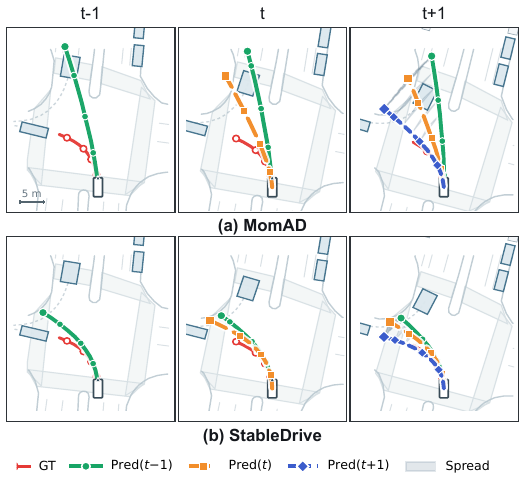}
\caption{
\textbf{Cross-Cycle Drift under a Shared Spatiotemporal Reference.}
Plans issued at $t-1$, $t$, and $t+1$ are transformed into each displayed current ego frame and restricted to their shared future timestamps.
MomAD~\cite{song2025momad} (top) shows increasing terminal dispersion across consecutive planning cycles, whereas StableDrive (bottom) keeps the aligned plans concentrated near the ground truth.
Gray shading indicates terminal dispersion; GT denotes ground truth.
}
\label{fig:temporal_consistency}
\end{figure}

\begin{figure*}[!t]
\centering
\includegraphics[width=0.98\textwidth]{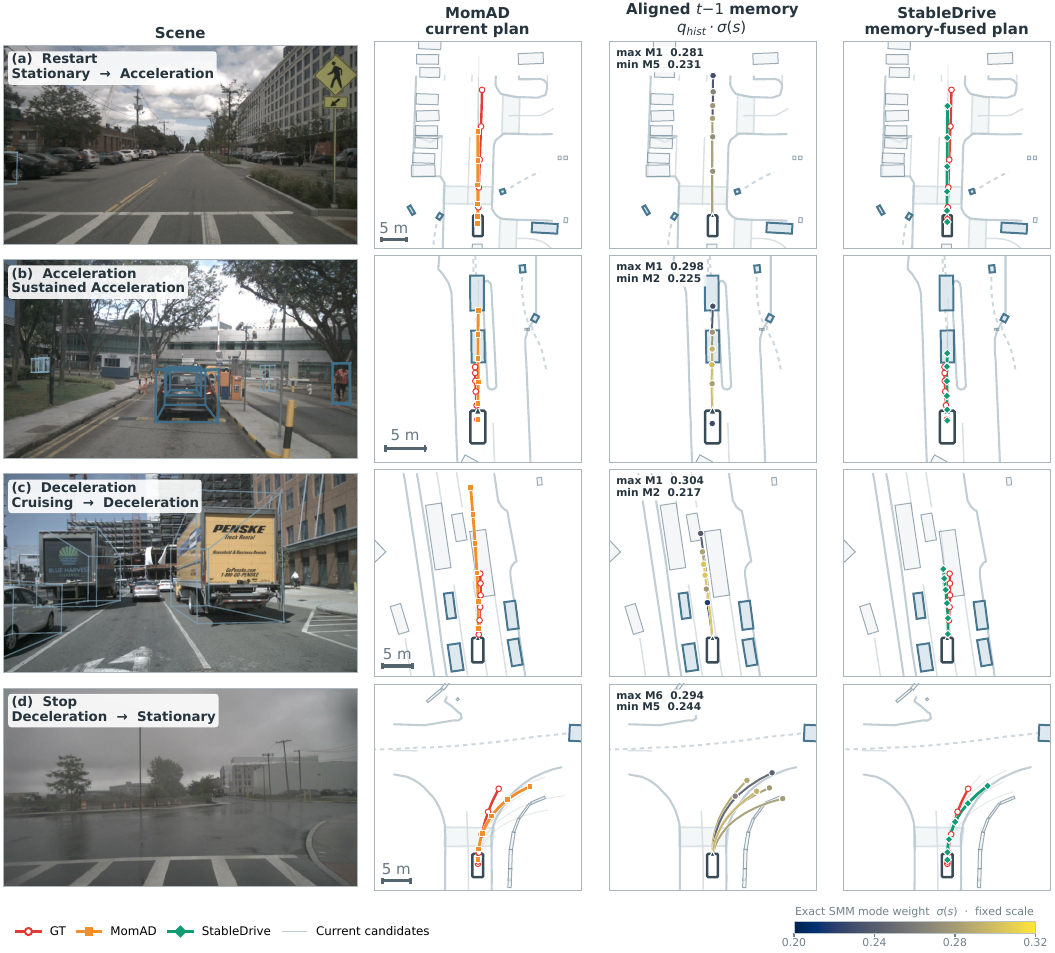}
\caption{
\textbf{Selective Reuse of One-Cycle History across Longitudinal Motion-Stage Transitions.}
Columns show the front-view scene, the current plan produced by MomAD~\cite{song2025momad}, pose- and time-aligned $t-1$ momentum modes colored by their exact SMM weights $\sigma(s)$, and the StableDrive memory-fused plan.
Rows show (a) restart, (b) sustained acceleration, (c) dense-traffic deceleration, and (d) stopping on a rainy curved road.
Memory colors use a fixed scale, and annotations identify the maximum- and minimum-weight modes.
The historical modes originate from the immediately preceding planning cycle, 0.5\,s earlier, and all BEV panels within each row use the same metric crop.
The non-uniform weights and the change in the dominant mode across the illustrated motion stages indicate that the contribution of one-cycle history is conditioned on the current planning state rather than applied uniformly across motion stages.
}
\label{fig:motion_stage_gallery}
\end{figure*}

\noindent\textbf{MSTS Retirement and Endpoint Construction.}
\Cref{tab:mstsretire} evaluates scaffold retirement and fixed-midpoint construction under equal training budgets. T1-on and T1-off share the same trained checkpoint and differ only in whether MSTS remains active at inference. Removing MSTS eliminates 0.811\,M active parameters, slightly improves L2 from 1.89 to 1.88\,m and TPC from 1.23 to 1.22\,m, and leaves the 1.35\% Col. Rate unchanged. The fixed midpoint further reduces L2, Col. Rate, and TPC to 1.83\,m, 1.19\%, and 1.20\,m, respectively. Because it operates directly on shared parameters, the final model preserves the SMM inference graph, with 87.153\,M active parameters and 192.728\,G FLOPs, and requires one checkpoint and one forward pass.

\begin{table*}[t]
\centering
\caption{Train--retire, endpoint-to-midpoint, and matched inference-cost analysis. Panel~A reports 4--6\,s averages under equal training budgets. T0 is the SMM-only endpoint; T1-on and T1-off use the same MSTS-trained checkpoint and differ only in whether MSTS is retained at inference. StableDrive is the fixed arithmetic midpoint ($\alpha=0.5$) of the architecture-aligned T0 and T1-off endpoints. Panel~B reports active parameters, FLOPs, latency, FPS, and peak allocated CUDA memory under steady-state, single-sample FP16 inference on one RTX~4090. T0, T1-off, and StableDrive share the same deployed SMM graph; their small throughput differences are therefore not ranked.}
\label{tab:mstsretire}
\begingroup
\scriptsize
\setlength{\tabcolsep}{4.0pt}
\renewcommand{\arraystretch}{1.10}
\noindent\textit{Panel A: Planning performance and construction role}\par
\vspace{0.15em}
\begin{tabularx}{\textwidth}{@{}p{1.45cm}p{2.60cm}*{6}{>{\centering\arraybackslash}X}@{}}
\toprule
Variant & Construction role & $\alpha$ & \shortstack{MSTS at\\inference} & \shortstack{Extra\\params} & \shortstack{L2 Avg.\\(m) $\downarrow$} & \shortstack{Col. Avg.\\(\%) $\downarrow$} & \shortstack{TPC Avg.\\(m) $\downarrow$}\\
\midrule
T0 & SMM endpoint & 0.0 & Off & 0 & 2.00 & 1.24 & 1.23\\
T1-on & Retention ablation & -- & On & +0.81M & 1.89 & 1.35 & 1.23\\
T1-off & MSTS endpoint & 1.0 & Off & 0 & 1.88 & 1.35 & 1.22\\
\rowcolor{gray!15}
\textbf{StableDrive} & \textbf{Fixed midpoint} & \textbf{0.5} & Off & 0 & \textbf{1.83} & \textbf{1.19} & \textbf{1.20}\\
\bottomrule
\end{tabularx}
\vspace{0.45em}
\noindent\textit{Panel B: Matched inference cost}\par
\vspace{0.15em}
\begin{tabularx}{\textwidth}{@{}>{\raggedright\arraybackslash}p{2.25cm}*{5}{>{\centering\arraybackslash}X}@{}}
\toprule
Variant & \shortstack{Active params\\(M) $\downarrow$} & \shortstack{FLOPs\\(G) $\downarrow$} & \shortstack{Latency\\(ms) $\downarrow$} & FPS $\uparrow$ & \shortstack{Peak memory\\(GB) $\downarrow$}\\
\midrule
T0 & 87.153 & 192.728 & 192.3 & 5.2 & 1.714\\
T1-on & 87.964 & 192.786 & 192.3 & 5.2 & 1.719\\
T1-off & 87.153 & 192.728 & 200.0 & 5.0 & 1.714\\
\rowcolor{gray!15}
\textbf{StableDrive} & 87.153 & 192.728 & 192.3 & 5.2 & 1.714\\
\bottomrule
\end{tabularx}
\endgroup
\end{table*}

\noindent\textbf{Training Dynamics.}
The SMM-only and MSTS-assisted branches in \Cref{fig:training_dynamics} exhibit closely aligned and bounded $\mathcal{L}_{\mathrm{base}}$ trajectories, while the motion-stage and longitudinal auxiliary losses settle to lower late-training levels. The curves show that MSTS provides additional supervision without destabilizing the shared planning objective; final checkpoint quality is evaluated separately in the endpoint and midpoint comparisons.

\noindent\textbf{History-Length Comparison.}
\label{sec:h4}
\Cref{tab:h4} compares one-, two-, and four-frame planning histories. The one-frame setting achieves the best average L2, Col. Rate, and TPC, with values of 1.20\,m, 0.66\%, and 0.85\,m, respectively. Extending the history window does not yield consistent gains and can introduce planning states that are no longer compatible with the current motion stage. The result supports selective use of recent history rather than indiscriminate temporal accumulation.

\begin{table*}[t]
\centering
\caption{History-length comparison of StableDrive using one, two, or four preceding planning cycles over the complete 1--6\,s horizon. L2 and Col. Rate are reported at each horizon and as six-horizon averages, while TPC is reported only as the six-horizon average. Lower is better for all metrics, and the best values are shown in bold.}
\label{tab:h4}
\begingroup
\scriptsize
\setlength{\tabcolsep}{1.5pt}
\renewcommand{\arraystretch}{1.08}
\begin{tabularx}{\textwidth}{@{}>{\raggedright\arraybackslash}p{2.65cm}*{14}{>{\centering\arraybackslash}X}>{\centering\arraybackslash}p{1.35cm}@{}}
\toprule
Variant & \multicolumn{7}{c}{L2 (m) $\downarrow$} & \multicolumn{7}{c}{Col. Rate (\%) $\downarrow$} & TPC (m) $\downarrow$\\
\cmidrule(lr){2-8}\cmidrule(lr){9-15}\cmidrule(l){16-16}
& 1~s & 2~s & 3~s & 4~s & 5~s & 6~s & Avg. & 1~s & 2~s & 3~s & 4~s & 5~s & 6~s & Avg. & Avg.\\
\midrule
\rowcolor{gray!15}
\textbf{StableDrive (1-frame )} & \textbf{0.28} & \textbf{0.55} & \textbf{0.89} & \textbf{1.32} & \textbf{1.80} & \textbf{2.35} & \textbf{1.20} & \textbf{0.01} & \textbf{0.09} & \textbf{0.30} & \textbf{0.65} & \textbf{1.16} & \textbf{1.78} & \textbf{0.66} & \textbf{0.85}\\
StableDrive (2-frame ) & 0.29 & 0.57 & 0.94 & 1.39 & 1.91 & 2.51 & 1.27 & 0.12 & 0.24 & 0.43 & 0.81 & 1.36 & 2.03 & 0.83 & 0.93\\
StableDrive (4-frame ) & 0.29 & 0.56 & 0.92 & 1.36 & 1.87 & 2.45 & 1.24 & 0.11 & 0.25 & 0.48 & 0.88 & 1.42 & 2.10 & 0.87 & 0.89\\
\bottomrule
\end{tabularx}
\endgroup
\end{table*}

\subsection{Qualitative Analysis}
\noindent\textbf{Cross-Cycle Planning Consistency.}
\Cref{fig:temporal_consistency} compares plans generated over three consecutive planning cycles under a shared spatiotemporal reference. After alignment to common future timestamps, MomAD~\cite{song2025momad} exhibits increasing terminal dispersion from $t-1$ to $t+1$, reflecting progressive cross-cycle planning drift. StableDrive keeps the aligned plans more concentrated around the ground-truth future path. This behavior is consistent with the lower TPC in the quantitative evaluation and supports the effect of selective historical reuse on planning consistency across successive cycles.

\noindent\textbf{Adaptation to Longitudinal Driving Transitions.}
\Cref{fig:motion_stage_gallery} examines one-cycle historical reuse in four representative cases: restarting, sustained acceleration, dense-traffic deceleration, and stopping on a rainy curved road. The history weights are non-uniform across the six cached planning modes, and the dominant mode varies across the illustrated motion stages. When the preceding motion trend remains compatible with the current driving and surrounding-traffic context, it provides a useful temporal prior; when that context requires a change in longitudinal motion, StableDrive adjusts the influence of the cached history before producing the updated plan. Together with the LT-nuScenes results, these cases indicate that reliable long-horizon planning benefits from retaining history that remains compatible with the evolving driving state rather than indiscriminately extending the history window.

\section{Conclusion}
We presented StableDrive, a long-horizon end-to-end planning framework that combines Selective Momentum Memory (SMM), Motion-Stage Training Scaffold (MSTS), and fixed midpoint checkpoint construction. Rather than treating historical planning information as uniformly beneficial, StableDrive selectively reuses compatible one-cycle history, while MSTS shapes motion-stage-aware long-horizon behavior during training and is removed before inference. The fixed midpoint construction then consolidates the complementary strengths of the two trained endpoints into a single checkpoint without changing the inference graph. Experiments on the full nuScenes validation set and the transition-focused LT-nuScenes subset show consistent improvements in trajectory accuracy, collision avoidance, and cross-cycle planning consistency over the complete 1--6\,s horizon. The history-length comparison, together with the transition-focused and qualitative analyses, further shows that simply extending the temporal window does not guarantee better planning, consistent with the stale-history failure mode highlighted throughout the study. Taken together, these findings suggest that reliable long-horizon planning depends less on retaining more history than on retaining history that remains compatible with the current motion stage.

\noindent\textbf{Limitations and Future Work.}
The current planner remains constrained by candidate generation and score-based selection, as a high-quality trajectory may be present in the candidate set yet remain unselected.
Future work will improve candidate generation and ranking while exploring more reliable selection strategies under the same single-checkpoint, single-forward inference design.
\bibliographystyle{IEEEtran}
\bibliography{references}

\begin{IEEEbiography}
[{\includegraphics[width=1in,height=1.25in,clip,keepaspectratio]{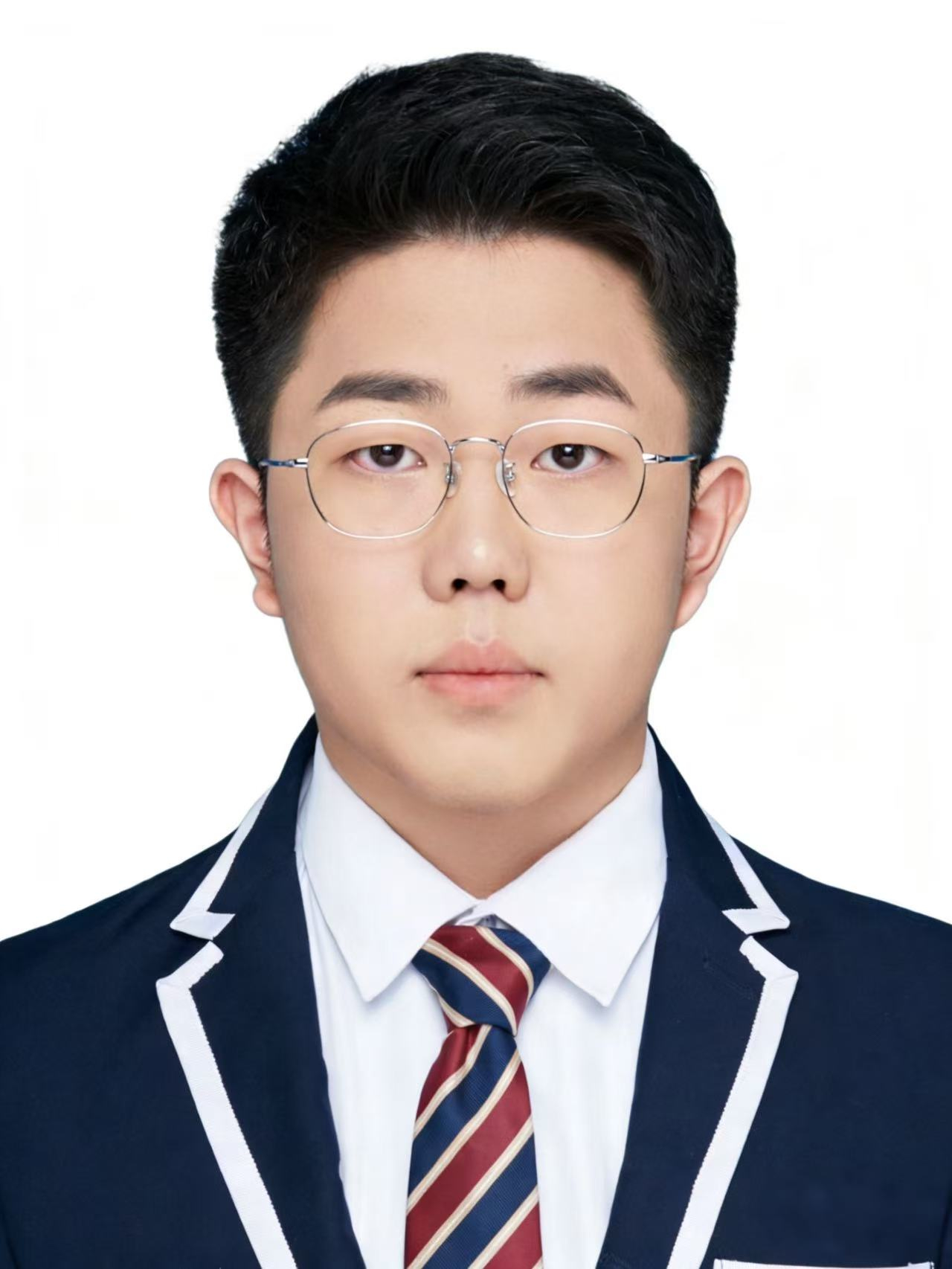}}]
{Yuchen Liu}
is currently a Research Assistant with Nanyang Technological University, Singapore. He is pursuing the B.E. degree in Internet of Things Engineering with the School of Computer Science and Technology, North University of China, Taiyuan, China. His research interests include end-to-end autonomous driving, long-horizon planning, world models, embodied intelligence, and vision-language-action models.
\end{IEEEbiography} \vspace{-2em}

\begin{IEEEbiography}[{\includegraphics[width=1in,height=1.25in,clip,keepaspectratio]{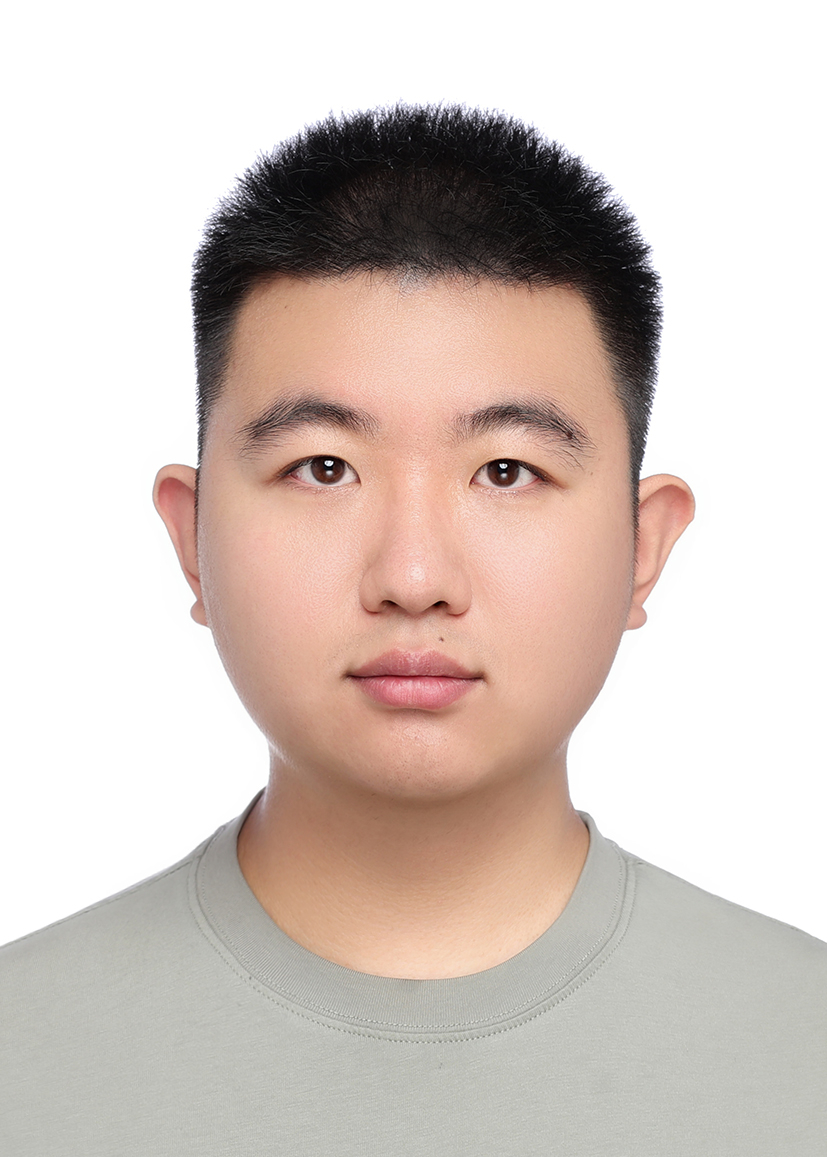}}]{Ziying Song}
 received the B.S. degree from Hebei Normal University of Science and Technology, China, in 2019, the M.S. degree from Hebei University of Science and Technology, China, in 2022, and the Ph.D. degree in computer science and technology from Beijing Jiaotong University, China, in March 2026. He is currently an Assistant Professor with the School of Artificial Intelligence, Yanshan University, China. His research interests include  autonomous driving, end-to-end autonomous driving, world models, embodied intelligence, and VLA models.
\end{IEEEbiography} \vspace{-2em}

\begin{IEEEbiography}[{\includegraphics[width=1in,height=1.25in,clip,keepaspectratio]{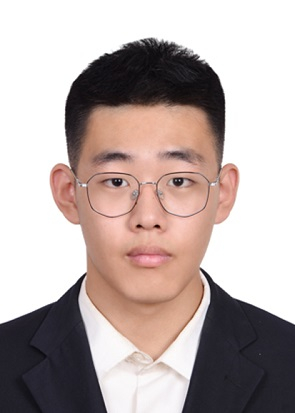}}]
{Shengkai Zhang}  studied for the B.S. degree in Computer Science and Technology at Hebei University, China, from 2022 to 2026. Since 2026, he has been pursuing the M.S. degree in Computer Science and Technology at Beijing Jiaotong University, China. His research interests include computer vision.
\end{IEEEbiography} \vspace{-2em}
\begin{IEEEbiography}[{\includegraphics[width=1in,height=1.25in,clip,keepaspectratio]{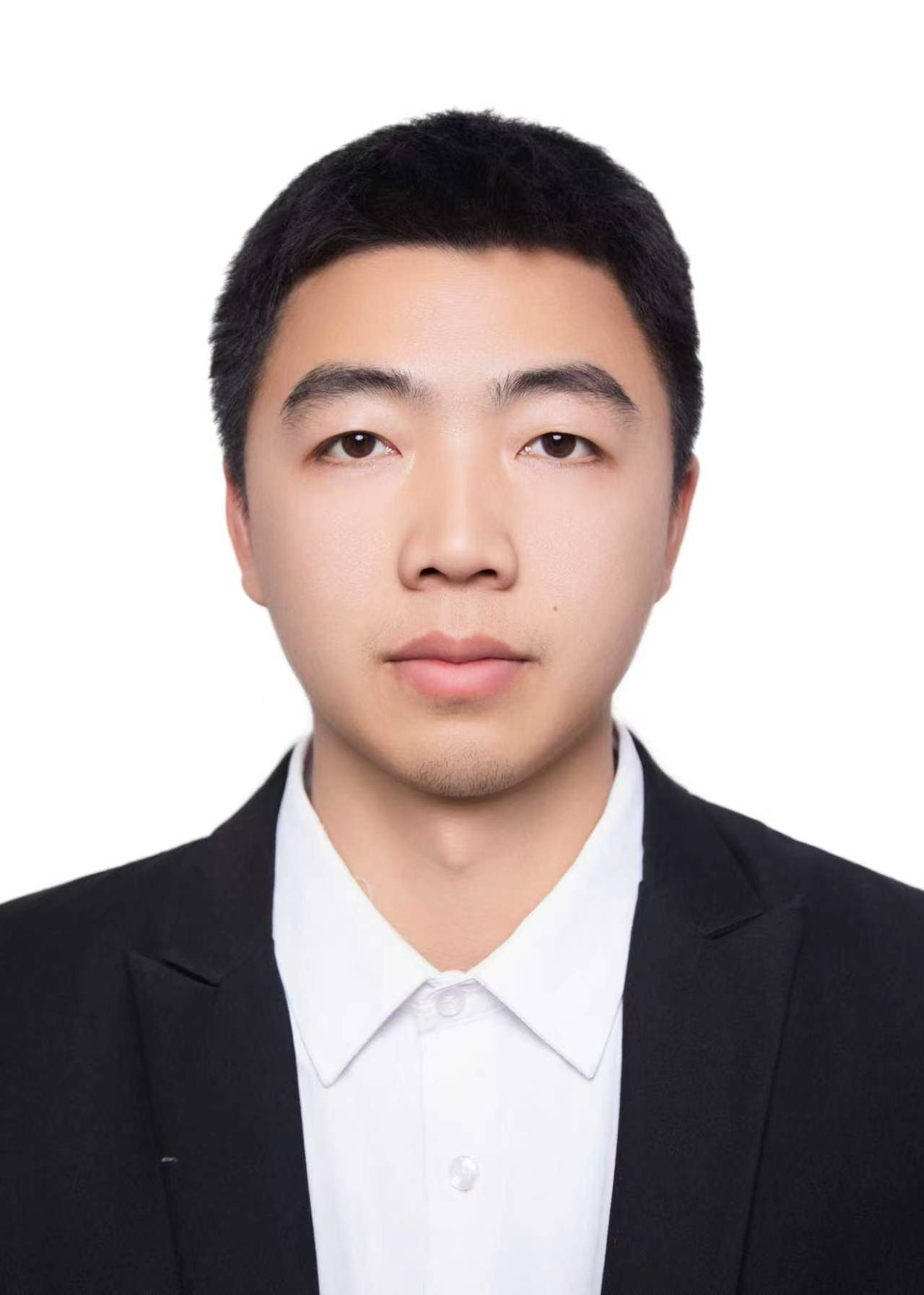}}] {Jiannan Chen} received the bachelor's degree in mechanical engineering and automation 
from the Beijing Union University, the master's and Ph.D. degrees in control theory and control engineering 
from Yanshan University. He is now a lecturer in the school of electrical engineering at Yanshan University. 
His research interests include EEG and EMG based human–computer interaction, deep learning, nonlinear 
control, and the control of UAVs.
\end{IEEEbiography} \vspace{-2em}
\begin{IEEEbiography}[{\includegraphics[width=1in,height=1.25in,clip,keepaspectratio]{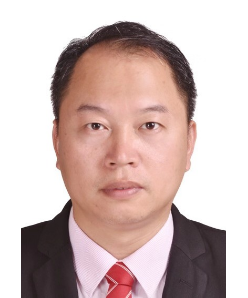}}]{Peiliang Wu} received the B.Sc. and Ph.D. degrees from Yanshan University, Qinhuangdao, China, in 2004 and 2010, respectively.  He is currently a Professor and Doctor Advisor with Yanshan University. He is a Member of the Academic Committee of Yanshan University, Member of the Standing Committee of the Youth Work Committee of the Chinese Artificial Intelligence Society, and the Vice Chairman of ACM Qinhuangdao. His research interests include robot learning and multi-agent systems.
\end{IEEEbiography} \vspace{-2em}

\begin{IEEEbiography}[{\includegraphics[width=1in,height=1.25in,clip,keepaspectratio]{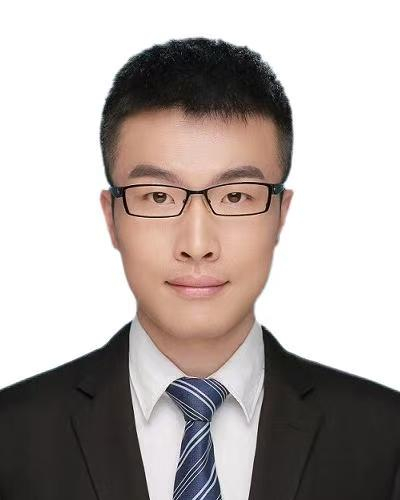}}]
{Lei Yang (Member, IEEE)} received the M.S. degree from the Robotics Institute, Beihang University, China, in 2018, and the Ph.D. degree from the School of Vehicle and Mobility, Tsinghua University, China, in 2024. From 2018 to 2020, he joined the Autonomous Driving R\&D Department of JD.COM as an algorithm researcher. Currently, he is a research fellow with the School of Mechanical and Aerospace Engineering, Nanyang Technological University, Singapore. His current research interests include autonomous driving, 3D scene understanding and world model.

\end{IEEEbiography} \vspace{-2em}

\begin{IEEEbiography}[{\includegraphics[width=1in,height=1.25in,clip,keepaspectratio]{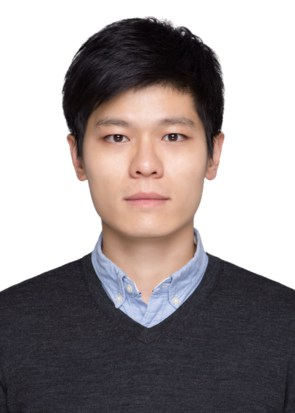}}] {Bin Sun} received the B.S. degree in Automotive Engineering from Jilin University, China, in 2014, and the M.S. degree from the Karlsruhe Institute of Technology (KIT), Germany, in 2017. From 2017 to 2022, he was with FAW-Volkswagen Automotive Co., Ltd. He received the Ph.D. degree in Automotive Engineering from Beihang University, China. He is currently with China Automotive Technology and Research Center Co., Ltd. (CATARC). His research interests include intelligent decision-making and planning strategies for autonomous vehicles, and end-to-end autonomous driving.
\end{IEEEbiography} \vspace{-2em}

\begin{IEEEbiography}[{\includegraphics[width=1in,height=1.25in,clip,keepaspectratio]{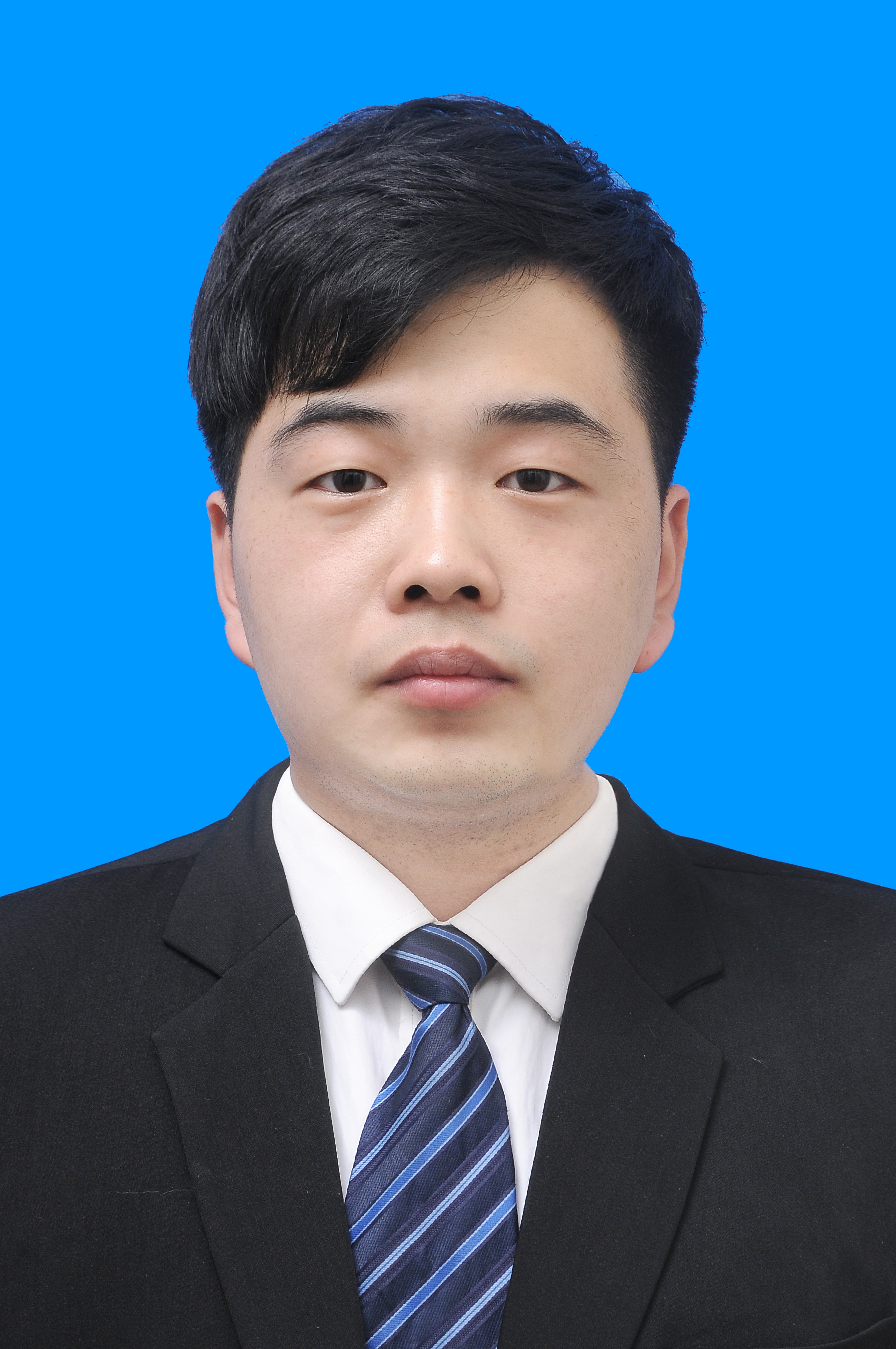}}]{Yan Gong} received a M.S. degree in Computer Science and Technology from Northeastern University, Shenyang, China, in 2023. From 2020 to 2023, he was a joint training master’s student with the School of Vehicle and Mobility at Tsinghua University, Beijing, China, where he also served as a research assistant. His research interests include autonomous driving, robotic perception, and multimodal fusion.
\end{IEEEbiography} \vspace{-2em}

\begin{IEEEbiography}[{\includegraphics[width=1in,height=1.25in,clip,keepaspectratio]{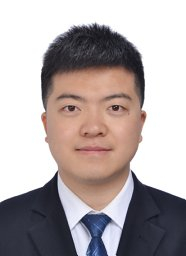}}] {Li Wang}
received his Ph.D. degree in mechatronic engineering at the State Key Laboratory of Robotics and System, Harbin Institute of Technology in 2020. He was a visiting scholar at Nanyang Technological University for two years. He was a postdoctoral fellow in the State Key Laboratory of Automotive Safety and Energy, and the School of Vehicle and Mobility, Tsinghua University. Currently, he is an assistant professor at School of Mechanical Engineering, Beijing Institute of Technology. His research interests include intelligent unmanned system, autonomous driving perception, and Multi-modal fusion.
\end{IEEEbiography} \vspace{-2em}
\end{document}